\documentclass[11pt]{article}

\usepackage[utf8]{inputenc}
\usepackage{amsmath, amssymb, amsthm}
\usepackage{graphicx}
\usepackage{hyperref}
\usepackage{geometry}
\usepackage{enumitem}
\usepackage{cite}
\usepackage{color}
\usepackage{booktabs}
\usepackage{microtype}
\usepackage{placeins}
\usepackage{float}
\usepackage{algorithm}
\usepackage{algpseudocode}
\usepackage{authblk}

\hypersetup{
  colorlinks=true,
  linkcolor=black,
  citecolor=black,
  urlcolor=black
}

\title{Graph In-Context Operator Networks for Generalizable Spatiotemporal Prediction}
\author{Chenghan Wu}
\author{Zongmin Yu}
\author{Boai Sun}
\author{Liu Yang\thanks{Corresponding author. Email: \texttt{yangliu@nus.edu.sg}}}
\affil{National University of Singapore}
\date{}

\begin{document}

\maketitle

\begin{abstract}
In-context operator learning enables neural networks to infer solution operators from contextual examples without weight updates. While prior work has demonstrated the effectiveness of this paradigm in leveraging vast datasets, a systematic comparison against single-operator learning using identical training data has been absent. We address this gap through controlled experiments comparing in-context operator learning against classical operator learning (single-operator models trained without contextual examples), under the same training steps and dataset. To enable this investigation on real-world spatiotemporal systems, we propose GICON (Graph In-Context Operator Network), combining graph message passing for geometric generalization with example-aware positional encoding for cardinality generalization. Experiments on air quality prediction across two Chinese regions show that in-context operator learning outperforms classical operator learning on complex tasks, generalizing across spatial domains and scaling robustly from few training examples to 100 at inference.
\end{abstract}

\section{Introduction}
\

The application of deep learning to solve partial differential equations (PDEs) has advanced substantially over the past decade. Early approaches focused on \emph{solution approximation}, where neural networks are trained to approximate the solution of a specific PDE instance. \cite{e2017deep} and \cite{han2018solving} pioneered deep learning for high-dimensional parabolic PDEs. The Deep Galerkin Method (DGM)~\cite{sirignano2018dgm} imposes constraints on neural networks to satisfy prescribed differential equations and boundary conditions. The Deep Ritz Method (DRM)~\cite{weinan2018deep} utilizes the variational form of PDEs to solve problems that can be transformed into equivalent energy minimization problems. PDE-Net~\cite{long2018pdenet} takes a complementary approach by learning PDE dynamics from data, enabling forward predictions using the inferred forward map. Physics-Informed Neural Networks (PINNs)~\cite{raissi2019physics} further advanced this paradigm by embedding PDE residuals directly into the loss function, enabling the network to satisfy both boundary conditions and governing equations. While these approaches achieve high accuracy, they require retraining for each new PDE instance, limiting their practical applicability.

Subsequent work shifted focus to \emph{operator learning}, which aims to learn the mapping from input functions (e.g., initial conditions, boundary conditions, or PDE coefficients) to solution functions. DeepONets~\cite{lu2021learning} and Fourier Neural Operators (FNOs)~\cite{li2021fourier} exemplify this paradigm, enabling a single trained model to generalize across different input functions for a fixed PDE family. However, these methods still require separate training for each distinct operator or PDE type.

In-context operator learning~\cite{yang2023context} extends this line of work by learning an \emph{operator learner}---a model capable of inferring new operators from prompted input-output function pairs (termed \emph{key}s and \emph{value}s, respectively\footnote{Prior ICON papers~\cite{yang2023context,yang2025fine,yang2024pde,cao2024vicon} use ``Cond'' and ``QoI'' for the input and output function of each example. We adopt the key-value terminology to align with the Transformer convention.}) without weight updates. This paradigm, inspired by in-context learning in large language models~\cite{radford2019language,brown2020language}, enables a single model to handle multiple PDE families by conditioning on contextual examples. Recent approaches such as ICON~\cite{yang2023context,yang2025fine,yang2024pde} and VICON~\cite{cao2024vicon} have shown promising results: \cite{yang2023context} demonstrated that increasing training diversity improves out-of-distribution generalization, \cite{yang2024pde} compared in-context operator learning against classical operator learning on 1D conservation laws and showed ICON's generalization to new PDE forms, and VICON~\cite{cao2024vicon} studied the effect of dataset combinations. However, existing comparisons between in-context and classical operator learning have two notable limitations: the two paradigms are trained on different datasets, compromising the fairness of the comparison, and the evaluations rely on synthetic data from simple PDEs rather than real-world observations. Consequently, it remains unclear whether in-context operator learning genuinely outperforms classical single-operator learning under a more controlled comparison, using identical training data and training steps. Additionally, extending these methods to real-world spatiotemporal physical systems introduces two practical challenges.

\begin{figure}[t]
\centering
\includegraphics[width=0.85\textwidth]{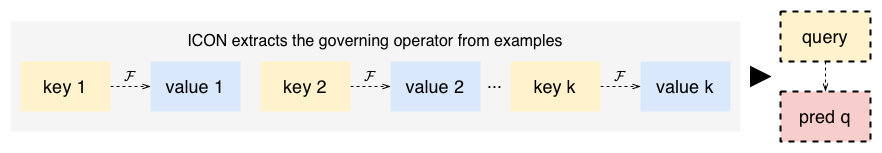}
\caption{Illustration of In-Context Operator Networks (ICON). Given $k$ contextual examples, each consisting of a key-value pair related by the governing operator $\mathcal{F}$, ICON infers the underlying operator from these in-context examples and applies it to map a new query input to its predicted output---all in a single forward pass without any weight updates.}
\label{fig:icon}
\end{figure}

First, the original ICON~\cite{yang2023context} and its variant ICON-LM~\cite{yang2025fine} represent each function by its sampled points, where each point corresponds to a token, leading to prohibitively long sequences for spatially dense data. VICON~\cite{cao2024vicon} addresses this through patchified image representations, which substantially reduce sequence length but implicitly assume regular grids and domain geometries. Many real-world physical systems---such as ground-based monitoring networks for air quality or meteorology---are inherently irregularly sampled and station-based.
Second, these models employ rigid positional encodings that tightly couple position embeddings to the number of in-context examples used during training, preventing models from leveraging additional examples at inference time.

\begin{figure}[t]
\centering
\includegraphics[width=\textwidth]{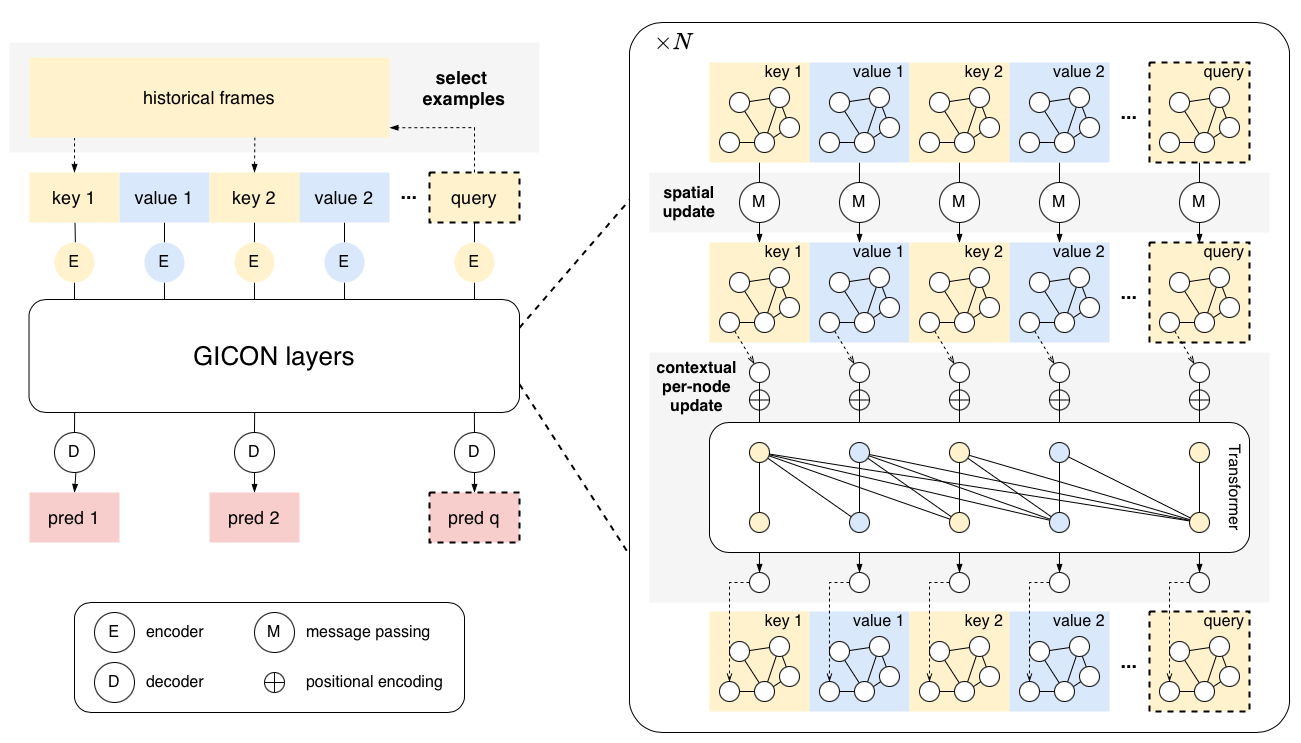}
\caption{Overview of the Graph In-Context Operator Network (GICON) architecture. \textbf{Left}: The overall pipeline selects contextual examples from historical frames, producing an interleaved sequence of keys and values along with the query. Each frame is processed by separate key and value encoders, passed through $N$ GICON layers, and decoded at key and query positions to produce predictions (during inference, only the query prediction is used). \textbf{Right}: Detailed structure of a single GICON layer, consisting of (1) \textit{spatial update} via message passing that aggregates information from neighboring nodes within each graph, and (2) \textit{contextual per-node update} using transformers that perform in-context learning across the example sequence with positional encodings.}
\label{fig:gicon_pipeline}
\end{figure}

To address these practical challenges and enable a systematic investigation of single-operator versus in-context operator learning, we propose GICON (Graph In-Context Operator Network), which introduces two architectural innovations:

\begin{enumerate}
    \item \textbf{Graph message passing for geometric generalization and scalability}: We replace patchified image representations with graph representations and decompose each GICON layer into a spatial update via message passing and a contextual per-node update via cross-example attention. Message passing encodes geometric structure, enabling generalization across different spatial domains; the decomposition avoids joint attention over all nodes and examples, providing scalability to large graphs.

    \item \textbf{Example-aware positional encoding for cardinality generalization}: We design a multi-level positional encoding scheme with two complementary components: (i) \emph{inter-example distinction} using example-aware attention biases to distinguish different contextual examples and the query, and (ii) \emph{key-value distinction} via learnable offsets to distinguish keys from values. By deriving positional information from content rather than fixed sequence indices, models trained with very few examples (e.g., 0--5) generalize stably to much larger numbers (up to 100) at inference time. 
\end{enumerate}

Additionally, GICON implements a retrieval-based approach for example selection, using FAISS-based similarity search to identify contextually relevant historical states as in-context examples.

Air quality prediction is a natural testbed for this investigation: as a dynamic spatiotemporal system, multiple operators can be constructed within a single dataset by varying the prediction horizon, providing operator diversity without requiring distinct PDE families. We validate GICON on air quality prediction tasks~\cite{wang2025knowairv2} across two Chinese regions: Beijing-Tianjin-Hebei and Yangtze River Delta. Our experiments reveal the following key findings:

\begin{itemize}
    \item \textbf{Geometric generalization}: Models trained on one region generalize effectively to another region with different graph structures and spatial geometries, suggesting that the learned representations are not tied to specific spatial configurations.

   \item \textbf{Example cardinality generalization}: As evidence of robust scaling, models trained with at most 5 examples maintain stable performance when evaluated with up to 100 examples, with models continuing to improve as example count increases.

    \item \textbf{In-context operator learning outperforms classical operator learning on complex tasks}: Under the same training steps and dataset, multi-operator in-context learning outperforms classical single-operator learning on complex operators, with performance generally improving as example count increases, even for out-of-distribution operators.

\end{itemize}

These findings demonstrate the effectiveness of in-context operator learning on real-world spatiotemporal systems and highlight the crucial role of operator diversity in enabling robust example utilization. Ablation studies further suggest that single-operator in-context learning may benefit from examples, though the improvement is limited and operator diversity remains essential for robust utilization. How examples are exploited in single-operator scenarios remains an open question. Our work provides new insights into the in-context operator learning paradigm and demonstrates GICON as an effective framework for extending it to irregularly sampled physical systems.

\section{Related Work}
\subsection{Operator Learning}
\

Operator learning~\cite{chen1995approximation,chen1995universal,lu2021learning,li2021fourier} addresses the challenge of approximating operators $\mathcal{G}: \mathcal{U} \to \mathcal{V}$, where $\mathcal{U}, \mathcal{V}$ are function spaces. These operators typically model solutions of physical systems and differential equations, representing the map from initial/boundary conditions and parameters to equation solutions.

DeepONets~\cite{lu2021learning,lu2022comprehensive} use branch and trunk networks to separately process inputs and query points, and have been extended to handle multiple input parametric functions in MIONet~\cite{jin2022mionet}. Fourier Neural Operators (FNOs)~\cite{li2021fourier,kovachki2023neural} leverage kernel integration approximations for PDE solutions by processing features in Fourier space, where kernel integrations can be efficiently computed through multiplications. These methods have been further extended to incorporate equation information~\cite{wang2021learning}, multiscale features~\cite{zhang2024bayesian,wen2022u}, different input/output meshes~\cite{zhang2023belnet,hao2023gnot}, complex geometries~\cite{li2023geometry}, and distributed learning~\cite{zhang2024d2no}. Applications of operator learning include weather prediction~\cite{kurth2023fourcastnet}, geosciences~\cite{jiang2024fourier}, biology and medicine~\cite{yin2024dimon}, and uncertainty quantification~\cite{moya2025conformalized}.

However, these approaches are typically designed to learn a single operator, requiring retraining when encountering different types of PDEs or time-step sizes. Pretraining-based approaches such as Poseidon~\cite{herde2024poseidon} and LeMON~\cite{sun2024lemon} address this through multi-operator pretraining followed by fine-tuning, improving sample efficiency on downstream tasks. Beyond fine-tuning, in-context approaches aim to generalize across operators without task-specific optimization.

\subsection{Graph Neural Networks for Operator Learning}
\

Graph Neural Networks (GNNs) have emerged as an effective framework for learning physics simulations by leveraging relational inductive biases~\cite{battaglia2018relationalinductivebiasesdeep}. Unlike grid-based methods, GNNs naturally handle irregular geometries and sparse interactions through message passing on graph structures, where nodes represent physical entities and edges encode their relationships.

GNNs have demonstrated strong performance in learning complex physical dynamics~\cite{sanchezgonzalez2020learningsimulatecomplexphysics}, including particle-based systems and fluid simulations. By iteratively aggregating information from neighboring nodes, GNNs can capture local interactions while maintaining computational efficiency. This approach has been extended to mesh-based simulations through MeshGraphNets~\cite{pfaff2021learningmeshbasedsimulationgraph}, enabling the learning of dynamics on adaptive meshes with varying resolutions. Furthermore, message-passing neural networks have been specifically designed as PDE solvers~\cite{brandstetter2022messagepassingneuralpde}, showing strong generalization capabilities across different equation parameters and domain geometries.

Building on these foundations, GNN-based neural operators have been developed to learn mappings between function spaces on irregular domains, including the Graph Neural Operator (GNO)~\cite{li2020neural}, Multipole GNO~\cite{li2020multipole}, Geometry-Informed Neural Operator (GINO)~\cite{li2023geometry}, and Region Interaction GNO (RIGNO)~\cite{mousavi2025rigno}. While these methods excel at handling irregular geometries, they typically learn a fixed operator for a specific PDE system. Combining the geometric flexibility of GNNs with the adaptability of in-context learning presents an opportunity to develop models that can handle both diverse geometries and multiple PDE types within a unified framework.

\subsection{In-Context Operator Networks}
\

In-Context Operator Networks (ICONs)~\cite{yang2023context} learn an \emph{operator learner} that infers new operators from prompted input-output pairs without weight updates. This enables generalization across diverse PDEs, ODEs, and mean-field control problems. Several extensions have advanced the core framework. ICON-LM~\cite{yang2025fine} incorporates a language-model-like autoregressive architecture with a ``next-function prediction'' training paradigm. \cite{yang2024pde} demonstrated generalization to new PDE forms unseen during training. \cite{zhang2025probabilistic} introduced GenICON, a probabilistic extension that enables generative modeling and uncertainty quantification by interpreting ICON as implicit Bayesian inference over solution operators.

The in-context operator learning paradigm has also been applied beyond differential equations. \cite{meng2025solving} applied ICON to optimal execution problems in finance, where the model infers price impact operators from few-shot trade trajectories. \cite{cole2026incontext} extended in-context operator learning to optimal transport on probability measure spaces. Theoretically, the paradigm is supported by robustness guarantees~\cite{liu2023does}, generalization bounds proving that task diversity is necessary and sufficient for out-of-domain generalization~\cite{cole2024incontext}, and connections to gradient descent in function spaces~\cite{mishra2025continuum}. Parallel approaches include PROSE~\cite{liu2024prose,liu2024proseFD}, which conditions on symbolic equation representations, \cite{chen2024dataefficient}, which combines unsupervised pretraining with in-context learning, and Zebra~\cite{serrano2025zebra} and ENMA~\cite{koupai2025enma}, which employ autoregressive latent-space generation for parametric PDEs.

Vanilla ICONs face quadratic computational complexity from point-wise function representation, limiting them to 1D problems or sparse 2D sampling. VICON~\cite{cao2024vicon} addresses this through patch-based sequences, enabling efficient dense 2D processing. However, existing methods share two practical limitations: grid-based representations that restrict applicability to irregular geometries, and fixed example counts during training that prevent cardinality generalization at inference.

\section{Method}
\subsection{Problem Setup}

\

We consider a forward problem defined on domain $\Omega \subseteq \mathbb{R}^d$ where $d$ is the spatial dimension ($d=2$ in this work), with state represented by $\mathbf{u}(\mathbf{x}, t) : \Omega \times [ 0, T ] \rightarrow \mathbb{R}^c$, where $c$ is the number of channels (e.g., temperature, pressure, velocity components). Our goal is to predict the future state $\mathbf{u}_{t+\Delta t}$ given historical observations.

\medskip
\textbf{Graph Representation.} To achieve better spatial symmetry and handle irregular spatial domains naturally, we represent the spatial domain $\Omega$ as a graph $\mathcal{G} = (\mathcal{V}, \mathcal{E})$, where $\mathcal{V}$ is the set of nodes and $\mathcal{E}$ is the set of edges. Each node $v_i \in \mathcal{V}$ corresponds to a spatial location $\mathbf{x}_i \in \Omega$, and edges $e_{ij} \in \mathcal{E}$ connect nodes based on spatial proximity or other physical connectivity criteria. This graph structure naturally accommodates irregular meshes, non-uniform spatial sampling, and complex boundary geometries that are common in real-world weather data. We denote the node features at time $t$ as $\mathbf{u}_t = \mathbf{u}(\cdot, t) \in \mathbb{R}^{|\mathcal{V}| \times c}$, where $|\mathcal{V}|$ is the number of nodes in the graph. Each row $\mathbf{u}_t^{(i)} \in \mathbb{R}^c$ represents the $c$-dimensional feature vector at node $v_i$.

\medskip
However, compared to VICON~\cite{cao2024vicon} which operates on dense pixel grids, our graph-based representation introduces a key challenge: the sparse discrete node sampling breaks the Markovian property. In dense continuous representations of first-order-in-time systems, a single snapshot at time $t$ captures sufficient local spatial information to infer the next state. In contrast, sparse graph nodes only observe the state at discrete locations, losing the fine-grained spatial structure between nodes. Consequently, a single snapshot $\mathbf{u}_t$ becomes insufficient to uniquely determine the future evolution. To recover the temporal dynamics, we must rely on a sequence of historical frames $\{\mathbf{u}_{t-\tau+1}, \ldots, \mathbf{u}_{t-1}, \mathbf{u}_{t}\}$ (where $\tau$ is the historical window size) from which the temporal evolution patterns can be inferred.

\medskip
\textbf{Operator Learning Framework.} We assume that the dynamics of the system can be modeled by a family of evolution operators $\{\mathcal{F}_{\Delta t}\}_{\Delta t > 0}$, such that
\begin{equation}
\mathbf{u}_{t+\Delta t} = \mathcal{F}_{\Delta t}(\mathbf{u}_{t-\tau+1}, \ldots, \mathbf{u}_{t-1}, \mathbf{u}_{t}).
\end{equation}
The operator $\mathcal{F}_{\Delta t}$ is inferred from a set of contextual examples that share the same time gap $\Delta t$. Unlike traditional numerical solvers that discretize governing equations, our approach learns the solution operator directly from observation data, enabling efficient predictions without requiring explicit knowledge of the underlying physics.

\medskip
\textbf{In-Context Learning Setup.} Following the in-context learning paradigm established in ICON~\cite{yang2023context, yang2025fine, yang2024pde}, we aim to develop a model that can adapt to new scenarios by leveraging a set of contextual examples. Each example consists of a pair $(\mathbf{k}, \mathbf{v})$, where $\mathbf{k}$ represents the key and $\mathbf{v}$ represents the value. In general, keys and values are input-output function pairs; in our spatiotemporal setting, keys correspond to historical observations and values to future states. In our graph-based setting, we denote $\mathbf{k}^{(j)} = \mathbf{u}^{(j)}_{t_j-\tau+1:t_j}$ and $\mathbf{v}^{(j)} = \mathbf{u}^{(j)}_{t_j+\Delta t}$, forming the example set $\mathcal{D} = \{(\mathbf{k}^{(j)}, \mathbf{v}^{(j)})\}_{j=1}^{N_d}$, where $N_d$ is the number of examples. These examples provide contextual information about the specific dynamics governing the current prediction task. The model processes both the examples and the query key $\mathbf{k}_{\text{query}} = \mathbf{u}_{t-\tau+1:t}$ to produce the prediction $\tilde{\mathbf{v}}_{\text{query}} = \hat{\mathbf{u}}_{t+\Delta t}$, allowing it to adapt its predictions based on the provided context without requiring gradient-based fine-tuning. This formulation performs ``next function prediction'' analogous to ``next token prediction'' in language models.

\medskip
\textbf{Training and Operator Consistency.} For efficient training parallelization, we employ a causal attention mask that exploits varying numbers of examples to compute all predictions in a single forward pass. The training loss is computed as the mean squared error (MSE) between predicted and ground truth values at both key positions and the query position. An important constraint is that all example pairs in the same sequence must be formed with the same operator mapping (i.e., the same $\Delta t$), ensuring the model learns a coherent operator from the context. However, operators can vary across different training sequences, enabling generalization across multiple operator types.

\subsection{Graph In-Context Operator Networks}

\subsubsection{Retrieval of Examples}

\

To effectively utilize the contextual examples in a scalable manner, we implement a retrieval mechanism that selects the most relevant examples from a potentially large pool of examples. This retrieval process serves two purposes: (1) it reduces computational costs by avoiding processing all available examples, and (2) it is expected to improve prediction quality by focusing on contextually relevant patterns.

\medskip
\textbf{Feature Extraction.} Given a query key $\mathbf{k}_{\text{query}} = \mathbf{u}_{t-\tau+1:t}$ and an example pool $\mathcal{D} = \{(\mathbf{k}^{(j)}, \mathbf{v}^{(j)})\}_{j=1}^{N_d}$, we first extract representative features from the key sequences. Rather than using all $\tau$ frames in the key, we focus on the last $\tau_r$ frames (where $\tau_r \leq \tau$) to capture the most recent dynamics. For each key sequence, we compute a pooled feature representation by aggregating spatial and temporal information. Specifically, for a key $\mathbf{k}$ consisting of frames $\{\mathbf{u}_{t-\tau_r+1}, \ldots, \mathbf{u}_{t}\}$, we apply global average pooling over the node dimension for each frame, resulting in temporal feature vectors, which are then concatenated or further aggregated to form a fixed-size representation $\mathbf{z} \in \mathbb{R}^d$.

\medskip
\textbf{Similarity-Based Retrieval.} We compute the similarity between the query feature $\mathbf{z}_{\text{query}}$ and each example feature $\mathbf{z}^{(j)}_{\text{ex}}$ using cosine similarity:
\begin{equation}
s_j = \frac{\mathbf{z}_{\text{query}}^\top \mathbf{z}^{(j)}_{\text{ex}}}{\|\mathbf{z}_{\text{query}}\|_2 \|\mathbf{z}^{(j)}_{\text{ex}}\|_2}.
\end{equation}
For efficient nearest neighbor search over large example pools, we employ the FAISS library~\cite{douze2025faiss}, which provides optimized implementations for approximate nearest neighbor search in high-dimensional spaces. FAISS enables us to scale to millions of examples while maintaining sub-linear query time through techniques such as product quantization and inverted file indexing.

\medskip
\textbf{Two-Stage Selection.} Based on the computed similarity scores $\{s_j\}_{j=1}^{N_d}$, we first select the top-$K$ examples with the highest scores:
\begin{equation}
\mathcal{D}_K = \{(\mathbf{k}^{(j)}, \mathbf{v}^{(j)}) : j \in \text{top-}K(\{s_j\})\},
\end{equation}
where $\text{top-}K(\cdot)$ returns the indices of the $K$ largest values. During training, to introduce randomness and prevent overfitting to specific example combinations, we randomly sample $k$ examples from $\mathcal{D}_K$ (where $k \leq K$) in each epoch to form the actual training context. This two-stage approach combines the benefits of similarity-based retrieval with stochastic data augmentation.

\subsubsection{Network Architecture}

\

Our Graph In-Context Operator Network (GICON) integrates graph neural network message passing with in-context operator learning through an architecture that processes sequences of graph-structured data. Given the sampled example set $\mathcal{D}_k = \{(\mathbf{k}^{(j)}, \mathbf{v}^{(j)})\}_{j=1}^{k}$ and the query $\mathbf{k}_{\text{query}}$, GICON processes these as an interleaved sequence to predict $\tilde{\mathbf{v}}_{\text{query}}$.

\medskip
\textbf{Input Projection and Sequence Construction.} Each key $\mathbf{k}$ and the query $\mathbf{k}_{\text{query}}$ consist of $\tau$ historical frames, which we flatten along the feature dimension and project to a common node embedding space via $\text{proj}_\mathbf{k}: \mathbb{R}^{\tau \times d_{\text{in}}} \to \mathbb{R}^{d_{\text{node}}}$. Each value $\mathbf{v}$ (a single frame) is projected via $\text{proj}_\mathbf{v}: \mathbb{R}^{d_{\text{in}}} \to \mathbb{R}^{d_{\text{node}}}$. We then construct an interleaved sequence by alternating keys and values: $[\mathbf{k}^{(1)}, \mathbf{v}^{(1)}, \mathbf{k}^{(2)}, \mathbf{v}^{(2)}, \ldots, \mathbf{k}^{(k)}, \mathbf{v}^{(k)}, \mathbf{k}_{\text{query}}]$, resulting in a sequence of length $2k+1$.

\medskip
\textbf{Graph Message Passing with In-Context Learning.} The core of GICON consists of $L$ stacked layers, each combining spatial message passing with temporal in-context learning. For each layer $\ell$, we maintain hidden states $\mathbf{h}^{(\ell)} \in \mathbb{R}^{(2k+1) \times |\mathcal{V}| \times d_{\text{node}}}$.

\medskip
Each GICON layer performs two sequential operations:
\begin{enumerate}
\item \textbf{Spatial Message Passing:} For each sequence position $t$ (in parallel across all positions), we aggregate information from neighboring nodes in the graph. Given hidden states $\mathbf{h}^{(\ell)}_{t} \in \mathbb{R}^{|\mathcal{V}| \times d_{\text{node}}}$ at sequence position $t$, we compute messages for each node $i$:
\begin{equation}
\mathbf{m}_{t,i} = \sum_{j \in \mathcal{N}(i)} \text{MLP}_{\text{msg}}([\mathbf{h}^{(\ell)}_{t,i}, \mathbf{h}^{(\ell)}_{t,j}, \mathbf{e}_{ij}])
\end{equation}
where $\mathcal{N}(i)$ denotes the neighbors of node $i$, and $\mathbf{e}_{ij}$ is the edge feature. The updated hidden states incorporate these aggregated messages: $\tilde{\mathbf{h}}^{(\ell)}_{t,i} = \mathbf{h}^{(\ell)}_{t,i} + \mathbf{m}_{t,i}$.

\item \textbf{Per-Node In-Context Learning:} For each node $i$ (in parallel across all nodes), we apply a transformer across the sequence dimension to enable in-context learning. We rearrange the hidden states to node-centric view $\tilde{\mathbf{h}}^{(\ell)}_{i} \in \mathbb{R}^{(2k+1) \times d_{\text{node}}}$ and apply:
\begin{equation}
\mathbf{h}^{(\ell+1)}_{i} = \text{Transformer}(\tilde{\mathbf{h}}^{(\ell)}_{i}, \text{mask}_{\text{causal}})
\end{equation}
where $\text{mask}_{\text{causal}}$ ensures autoregressive prediction by preventing positions from attending to future positions in the sequence.
\end{enumerate}

\medskip
This design enables each node to learn from spatial neighbors while simultaneously performing in-context learning across contextual examples, combining the geometric flexibility of GNNs with the adaptability of in-context learning.

\medskip
\textbf{Positional Encoding for Example Distinction.} A critical challenge in GICON is distinguishing between (1) different contextual examples versus the query, and (2) keys versus values within the sequence. We introduce two complementary positional encoding strategies:

For inter-example distinction, we introduce an example-aware attention bias $\mathbf{A} \in \mathbb{R}^{H \times (2k+1) \times (2k+1)}$ that operates at the example level. We first extract key tokens (at even sequence positions) and pool them across nodes to obtain example-level representations $\bar{\mathbf{H}}_\mathbf{k} \in \mathbb{R}^{(k+1) \times d_{\text{node}}}$. These are projected through an MLP to yield example embeddings $\mathbf{Z} = \text{MLP}(\bar{\mathbf{H}}_\mathbf{k}) \in \mathbb{R}^{(k+1) \times d_{\text{node}}}$, from which we compute head-specific pairwise similarities:
\begin{equation}
\mathbf{S} = \frac{(\mathbf{W}_q \mathbf{Z})(\mathbf{W}_k \mathbf{Z})^\top}{\sqrt{d_{\text{node}}/H}} \in \mathbb{R}^{H \times (k+1) \times (k+1)}
\end{equation}
The attention bias for each position pair $(i,j)$ is then set as $\mathbf{A}[:,i,j] = \mathbf{S}[:,\lfloor i/2 \rfloor, \lfloor j/2 \rfloor]$, so that each value token inherits the bias of its corresponding key within the same example. This content-aware bias encourages attending to examples with similar keys, and since it is derived from content rather than positional indices, it naturally generalizes to any number of examples at inference time.

For key-value distinction, we add learnable offset vectors $\pm \mathbf{r}$ to the token embeddings, where key tokens receive $+\mathbf{r}$ and value tokens receive $-\mathbf{r}$, creating a separation between keys and values.

This multi-level positional encoding scheme enables GICON to effectively leverage the hierarchical structure of in-context operator learning on graphs, distinguishing both the example-query relationship and the key-value pairing within each example. We provide detailed pseudo-code in Algorithm~\ref{alg:pos-encoding} (Appendix~\ref{sec:appendix-pos-encoding}).

\section{Experiments}
\

We conduct experiments to evaluate whether in-context operator learning can outperform classical single-operator learning on complex tasks, and examine how operator diversity affects the utilization of in-context examples. We further validate GICON's generalization capabilities across example counts and spatial domains. All experiments are performed on two air quality datasets with consistent qualitative trends.

\subsection{Experimental Setup}

\subsubsection{Datasets}
\

We evaluate on air quality monitoring datasets from two major Chinese regions, following the benchmark established by \cite{wang2025pcdcnet}. The BTHSA dataset covers Beijing-Tianjin-Hebei and Surrounding Areas ($\sim$430,000 $\text{km}^2$, 28 cities) with 228 monitoring stations, while the YRD dataset covers the Yangtze River Delta ($\sim$270,000 $\text{km}^2$) with 127 stations. Together, these provide 70,128 hours of observations spanning 2016--2023.

Each station records 10 raw features: 2 air quality variables (hourly PM$_{2.5}$ and O$_3$ concentrations from CNEMC) and 8 meteorological variables (wind components, temperature, precipitation, and surface pressure from ERA5 reanalysis). We derive 3 additional features: hour of day as a temporal encoding, and wind speed/direction computed from wind components ($u_{100}$, $v_{100}$) using MetPy~\cite{metpy}, yielding 13 input features per station. Spatial connections between stations are established using a 200km geodesic distance threshold; edges are further filtered by removing connections where intervening terrain exceeds 1200m above source or destination altitude, modeling physical barriers to pollutant transport. Each edge carries attributes of geodesic distance and direction. We use 2016--2022 as the example retrieval pool, 2017--2022 for training, and 2023 for testing. Both datasets are publicly available.\footnote{\url{https://zenodo.org/records/15614907}}

\subsubsection{Training and Evaluation Protocol}
\

Each operator corresponds to temporal evolution with time gap $\Delta t$, and we set the historical window size $\tau = \tau_r = 24$ (i.e., each key consists of 24 hourly frames). The classical single-operator learning baseline trains on a fixed $\Delta t \in \{1, 4, 12, 24\}$ hours without contextual examples ($k = 0$). For the ICON paradigm, we primarily use multi-operator training with $\Delta t$ sampled uniformly from $[1, 24]$ hours, with maximum example counts $k \in \{1, 2, 5\}$; we additionally examine single-operator in-context learning ($k \in \{1, 2, 5\}$) as an ablation study (Section~\ref{sec:ablation}). All models are trained for the same number of steps (90,000), but since each valid $\Delta t$ value generates a separate set of training sequences, single-operator models see each sample $\sim$$24\times$ more often during training, while multi-operator models trade per-sample repetition for operator diversity. Models are evaluated with up to 100 examples to assess cardinality generalization. Note that all models, including single-operator models and those trained without contextual examples ($k = 0$), are provided with examples during evaluation; this allows us to test whether models trained on a single operator or without examples can still leverage in-context examples at inference. We adopt Root Mean Square Error (RMSE) as our primary evaluation metric, computed over all stations as $\text{RMSE} = \sqrt{\frac{1}{N}\sum_{i=1}^{N}(\hat{y}_i - y_i)^2}$. Detailed model architecture, training objective and optimization configuration are provided in Appendix~\ref{sec:appendix-training-config}.

\subsection{Example Cardinality Generalization}
\label{sec:cardinality}
\

\begin{figure}[t]
    \centering
    \begin{minipage}{0.32\textwidth}
        \centering
        \includegraphics[width=\linewidth]{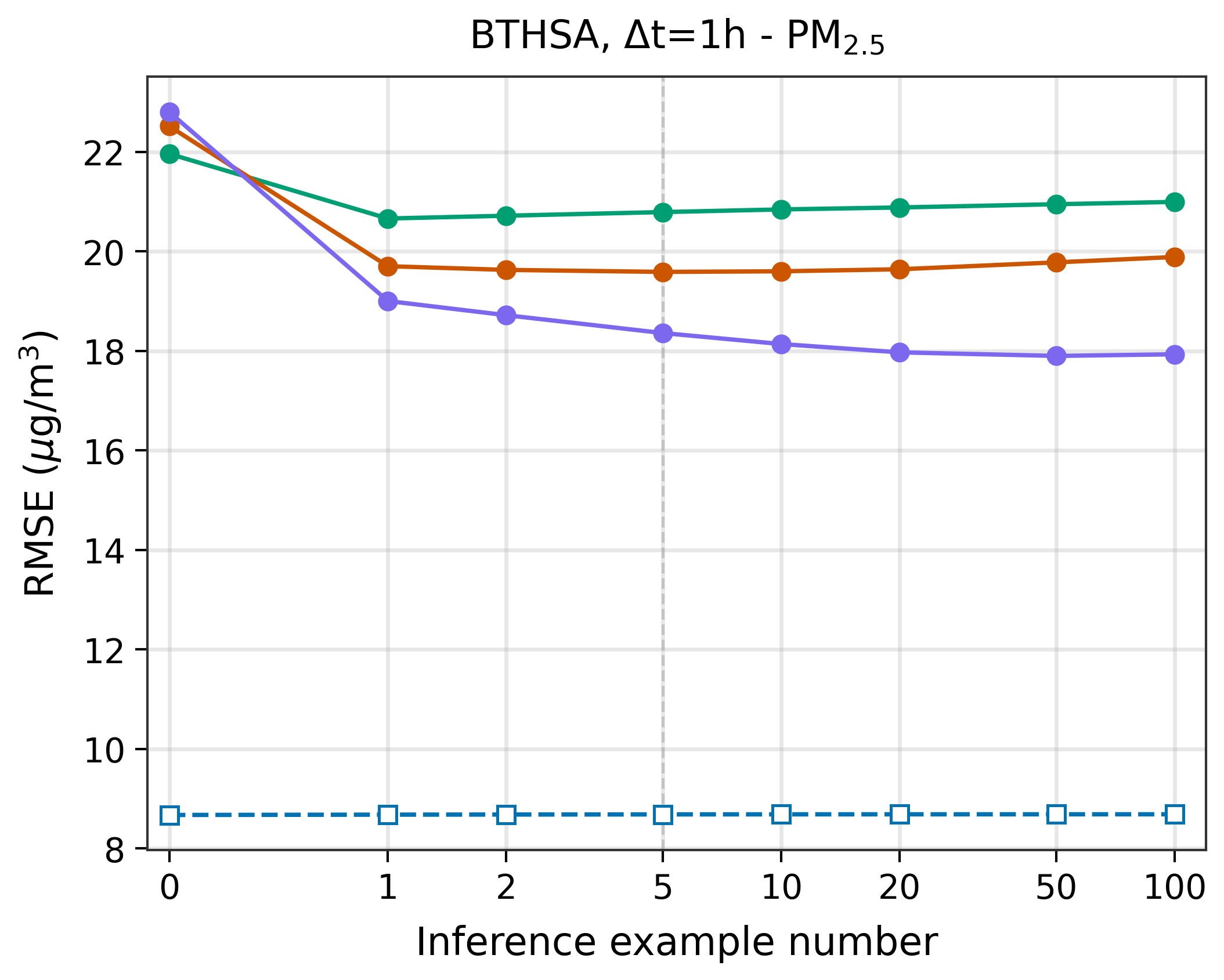}
    \end{minipage}
    \hfill
    \begin{minipage}{0.32\textwidth}
        \centering
        \includegraphics[width=\linewidth]{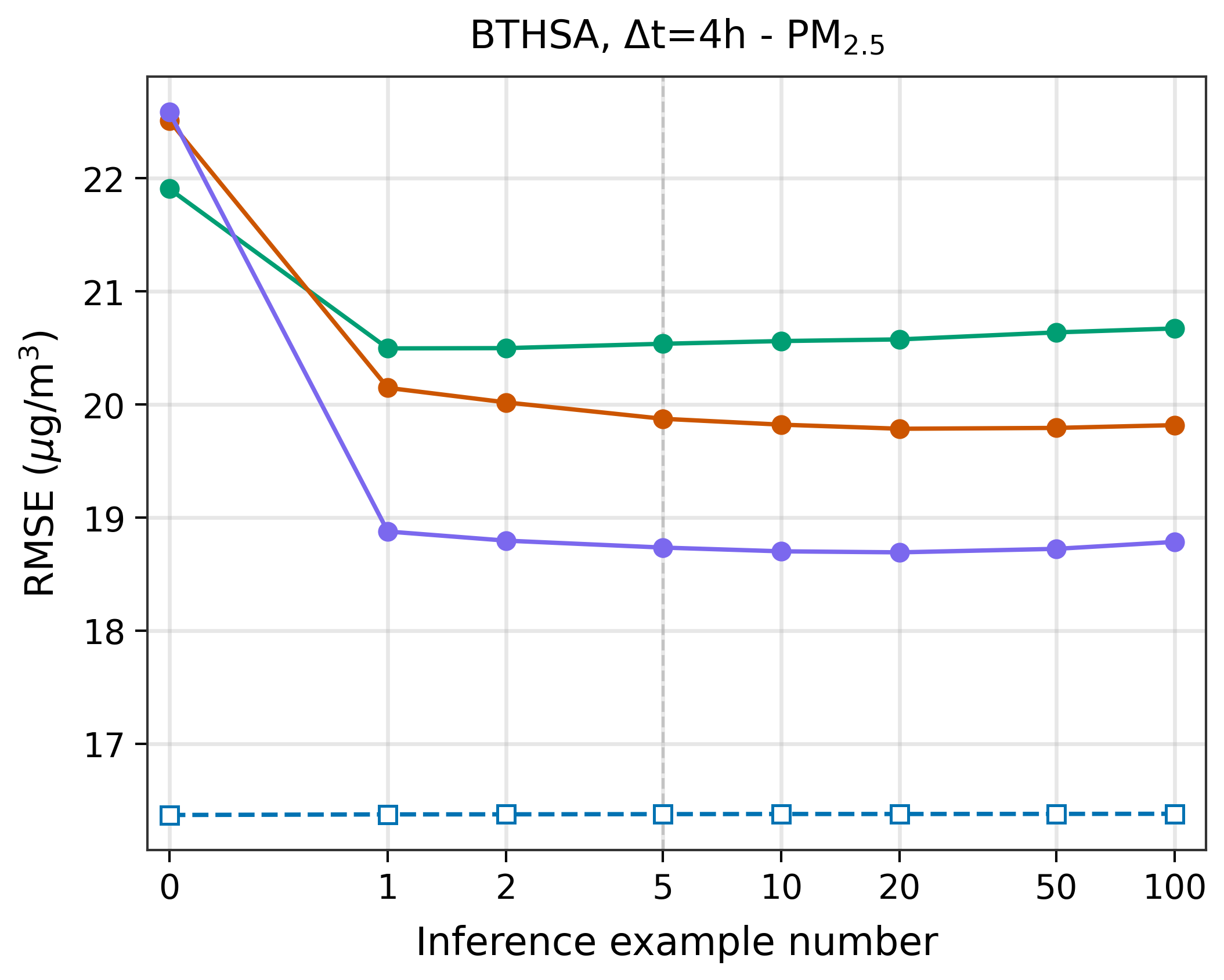}
    \end{minipage}
    \hfill
    \begin{minipage}{0.32\textwidth}
        \centering
        \includegraphics[width=\linewidth]{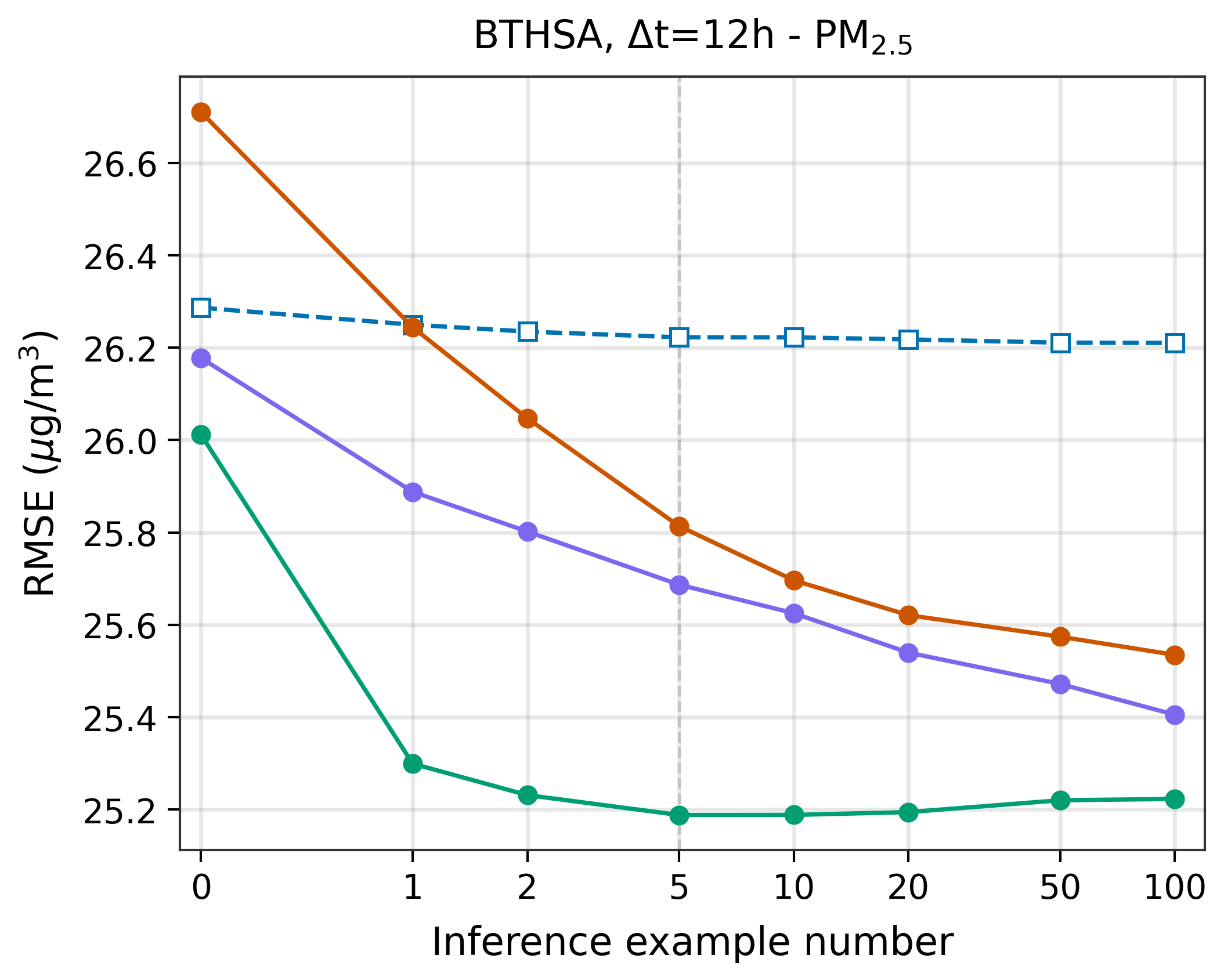}
    \end{minipage}
    \\[0.5em]
    \begin{minipage}{0.32\textwidth}
        \centering
        \includegraphics[width=\linewidth]{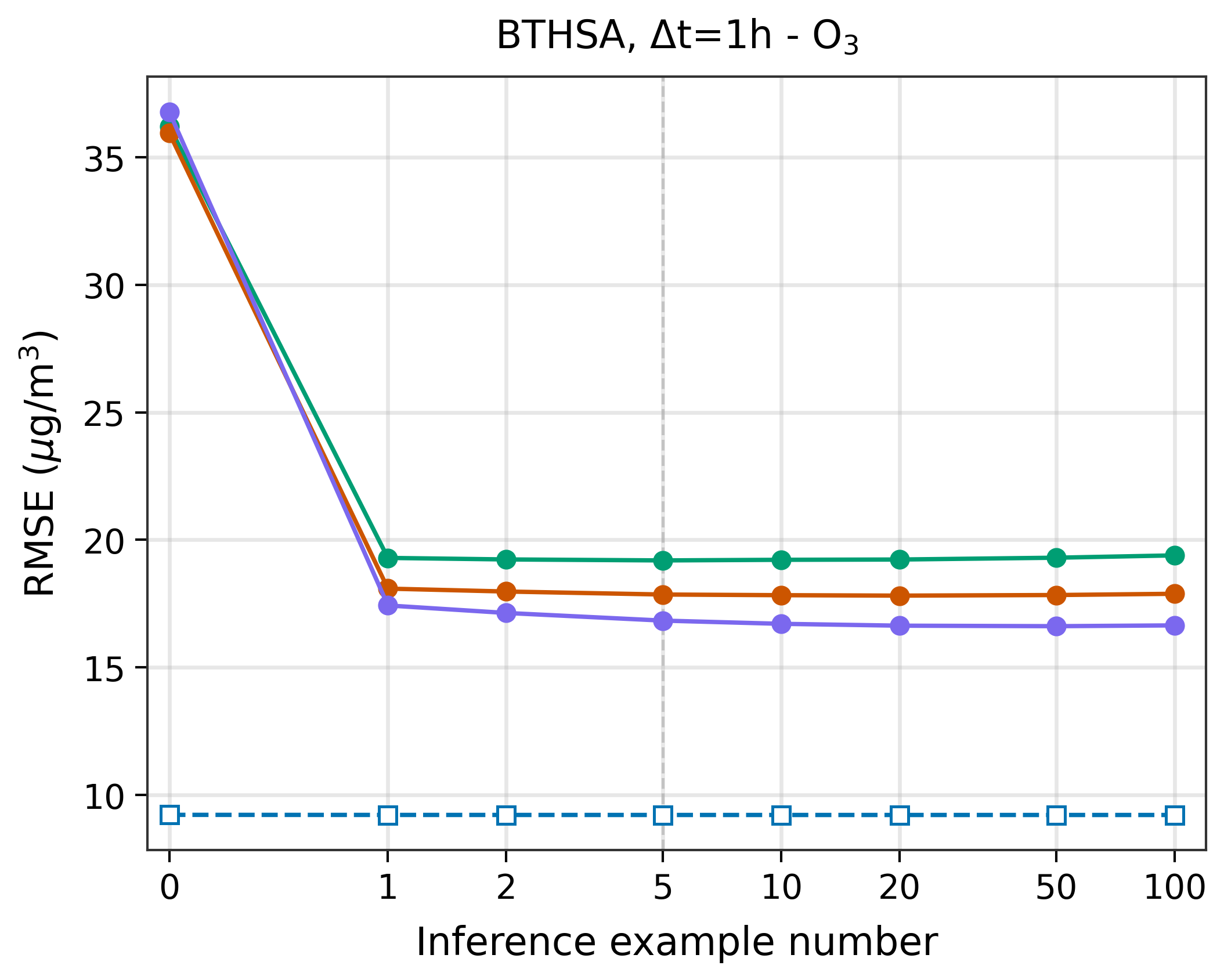}
    \end{minipage}
    \hfill
    \begin{minipage}{0.32\textwidth}
        \centering
        \includegraphics[width=\linewidth]{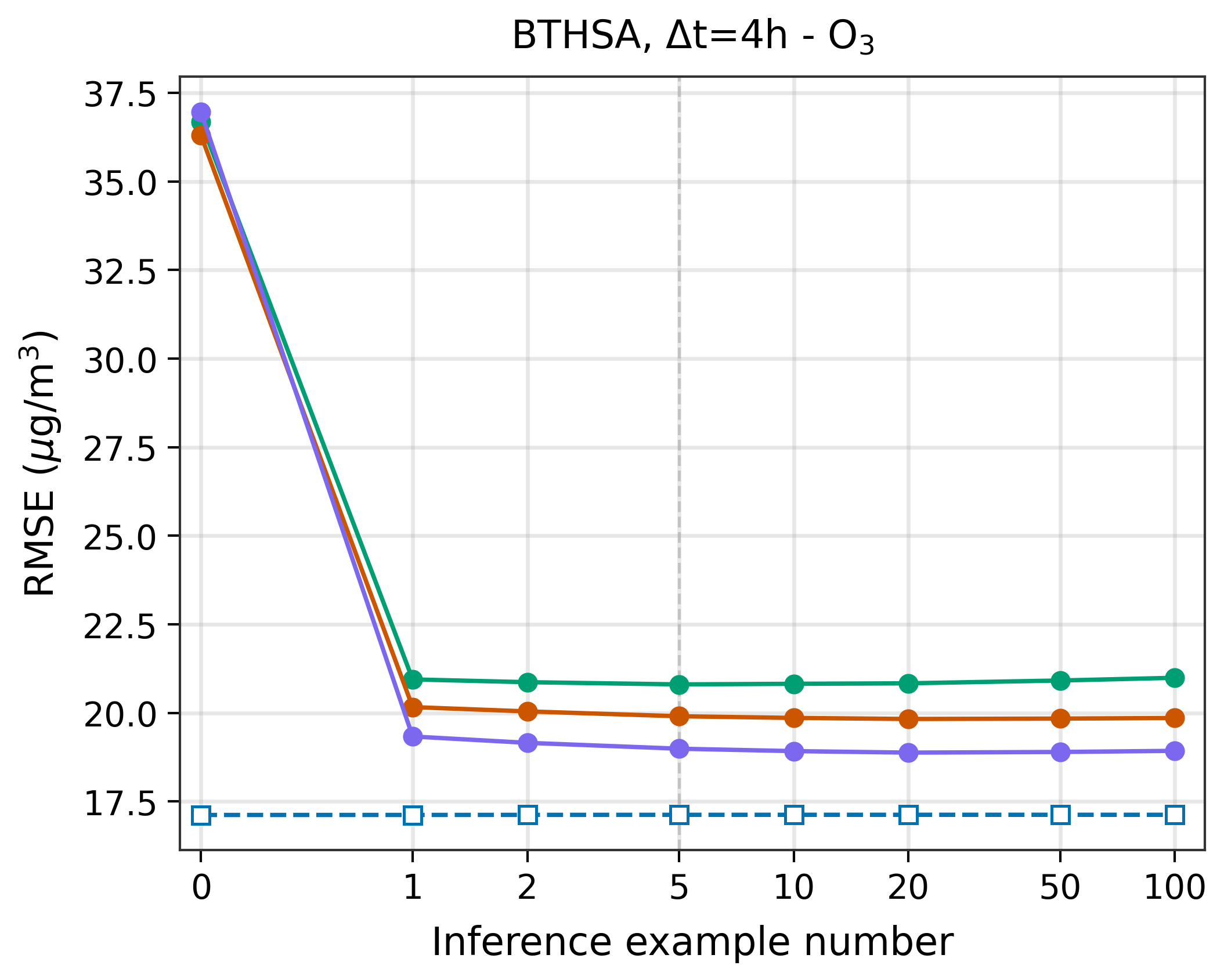}
    \end{minipage}
    \hfill
    \begin{minipage}{0.32\textwidth}
        \centering
        \includegraphics[width=\linewidth]{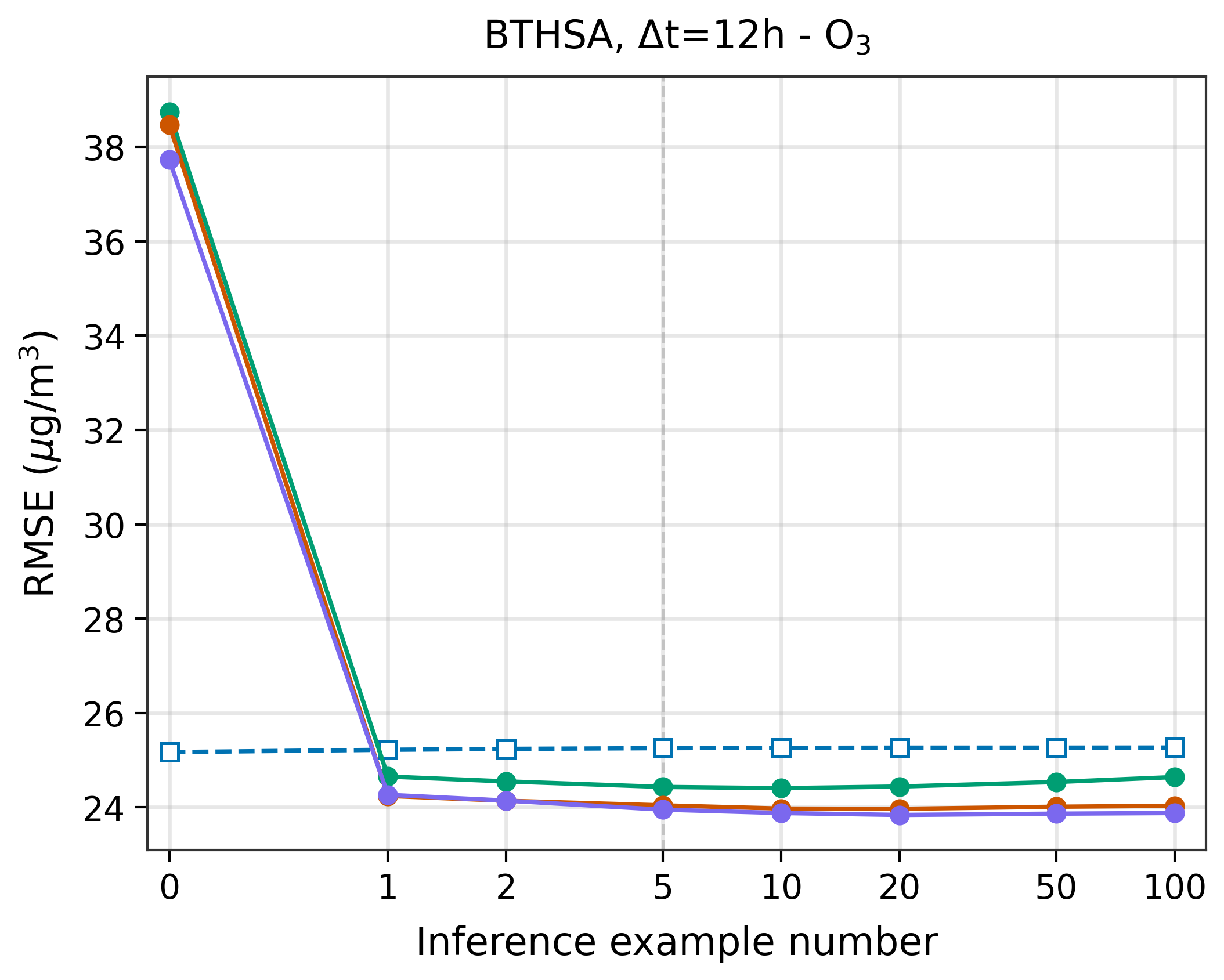}
    \end{minipage}
    \\[0.3em]
    \includegraphics[width=0.7\textwidth]{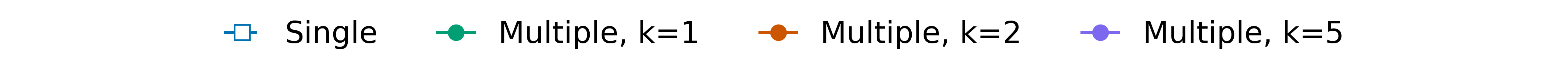}
    \caption{Example cardinality generalization on BTHSA for simple to moderate operators. Top: PM$_{2.5}$. Bottom: O$_3$. Left to right: $\Delta t = 1, 4, 12$h. Classical single-operator learning achieves lower RMSE for simple operators ($\Delta t = 1, 4$h), while ICON with operator diversity outperforms the baseline at $\Delta t = 12$h given sufficient examples. All models are evaluated with up to 100 examples despite training with at most 5.}
    \label{fig:demo_cardinality}
\end{figure}

\begin{figure}[t]
    \centering
    \begin{minipage}{0.48\textwidth}
        \centering
        \includegraphics[width=\linewidth]{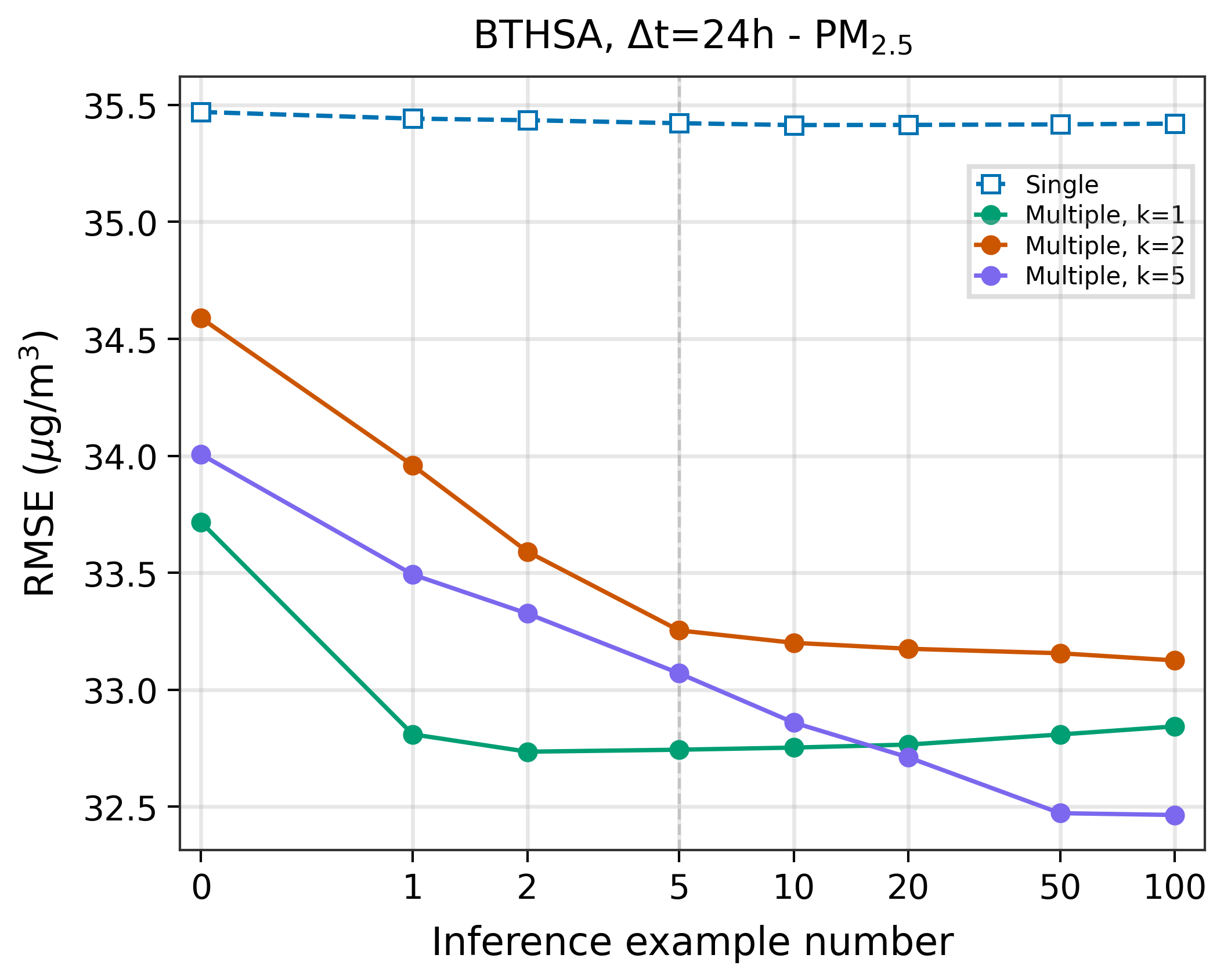}
    \end{minipage}
    \hfill
    \begin{minipage}{0.48\textwidth}
        \centering
        \includegraphics[width=\linewidth]{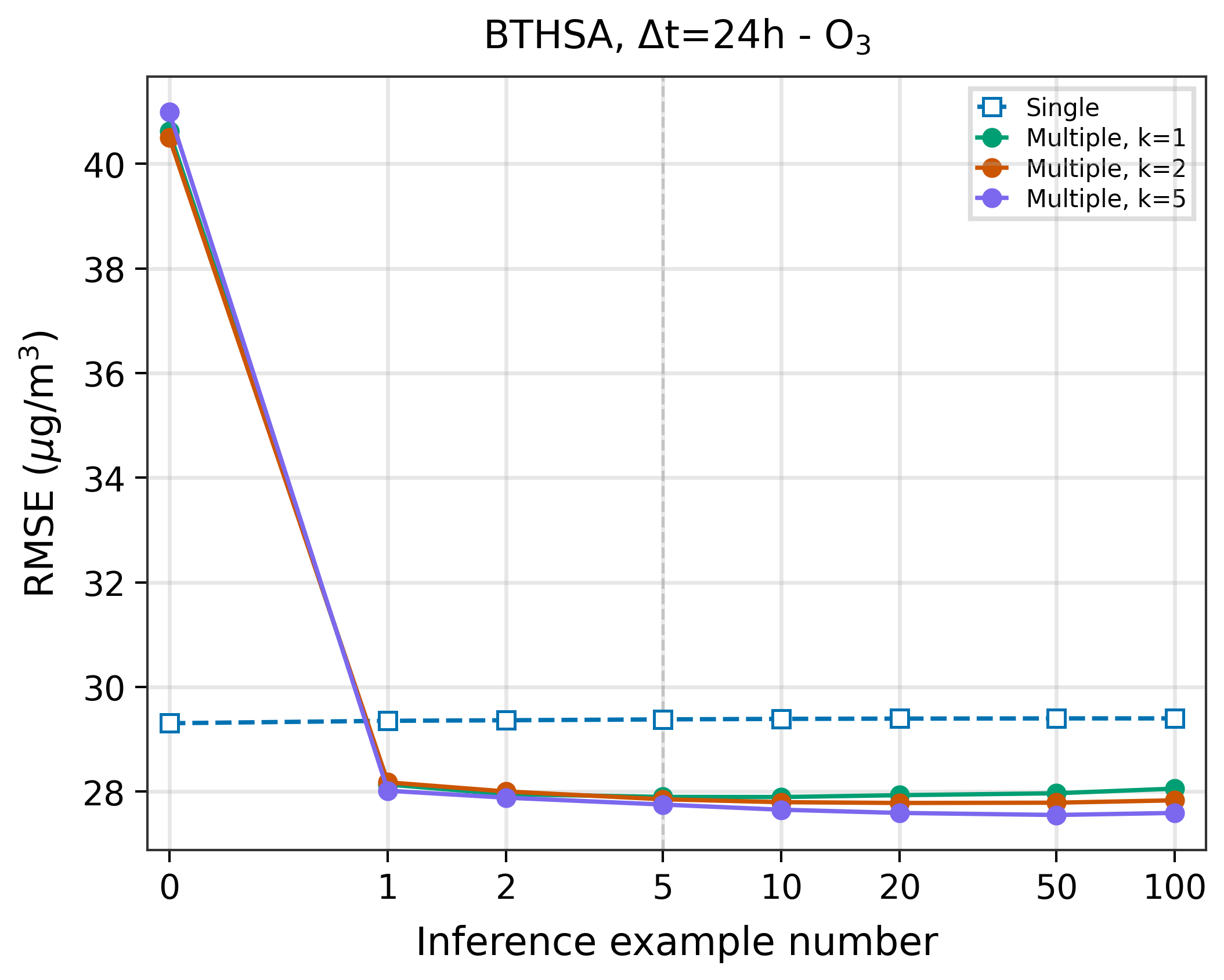}
    \end{minipage}
    \caption{Example cardinality generalization on BTHSA at $\Delta t = 24$h. Left: PM$_{2.5}$. Right: O$_3$. For this complex operator, ICON with operator diversity surpasses the single-operator baseline with sufficient examples, with error decreasing for PM$_{2.5}$ and a sharp initial drop followed by stable performance for O$_3$.}
    \label{fig:demo_cardinality_24}
\end{figure}

We examine how example cardinality generalization varies with operator complexity by evaluating models across $\Delta t \in \{1, 4, 12, 24\}$h (Figures~\ref{fig:demo_cardinality} and~\ref{fig:demo_cardinality_24}). Models trained with maximum example count $k \in \{1, 2, 5\}$ maintain stable performance when evaluated with up to 100 examples across all $\Delta t$ values, confirming that the example-aware positional encoding generalizes beyond the training example count.

For simple operators ($\Delta t = 1, 4$h), classical single-operator learning achieves lower RMSE than ICON with operator diversity, as the task is simple enough that a dedicated model can specialize effectively. As operator complexity increases ($\Delta t = 12, 24$h), ICON with operator diversity surpasses the single-operator baseline with sufficient examples, with the advantage becoming more pronounced at larger $\Delta t$. The contrast is especially clear for O$_3$, where ICON with operator diversity exhibits a substantial RMSE drop with just one example, while the single-operator baseline remains flat regardless of example count.

These results suggest that examples provide generally increasing benefit when operator diversity is present and the task is sufficiently complex. Additional results for the YRD dataset are provided in Appendix~\ref{sec:appendix-demo-cardinality}.

\subsection{Out-of-Distribution Operator Extrapolation}
\

\begin{figure}[!b]
    \centering
    \begin{minipage}{0.48\textwidth}
        \centering
        \includegraphics[width=\linewidth]{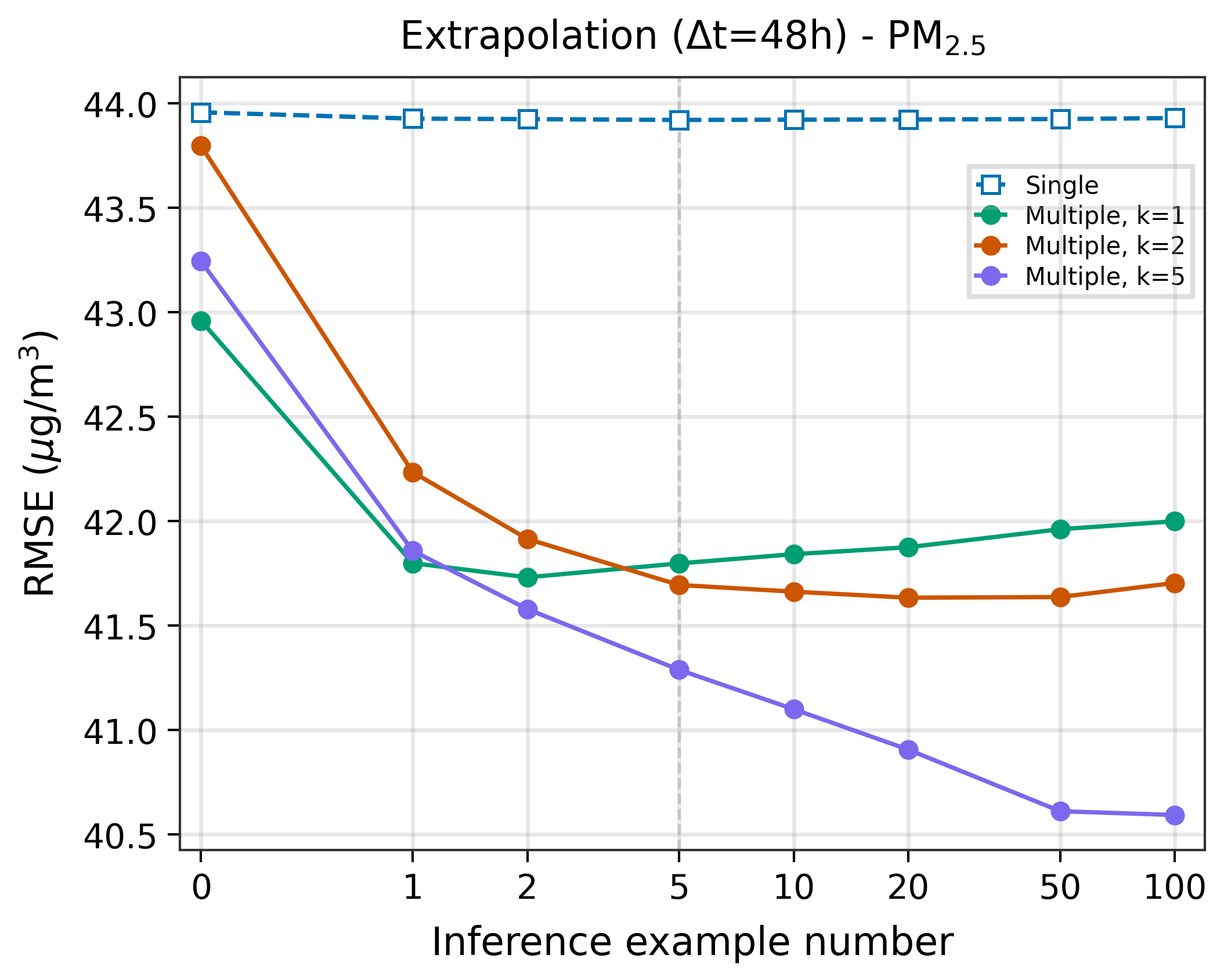}
    \end{minipage}
    \hfill
    \begin{minipage}{0.48\textwidth}
        \centering
        \includegraphics[width=\linewidth]{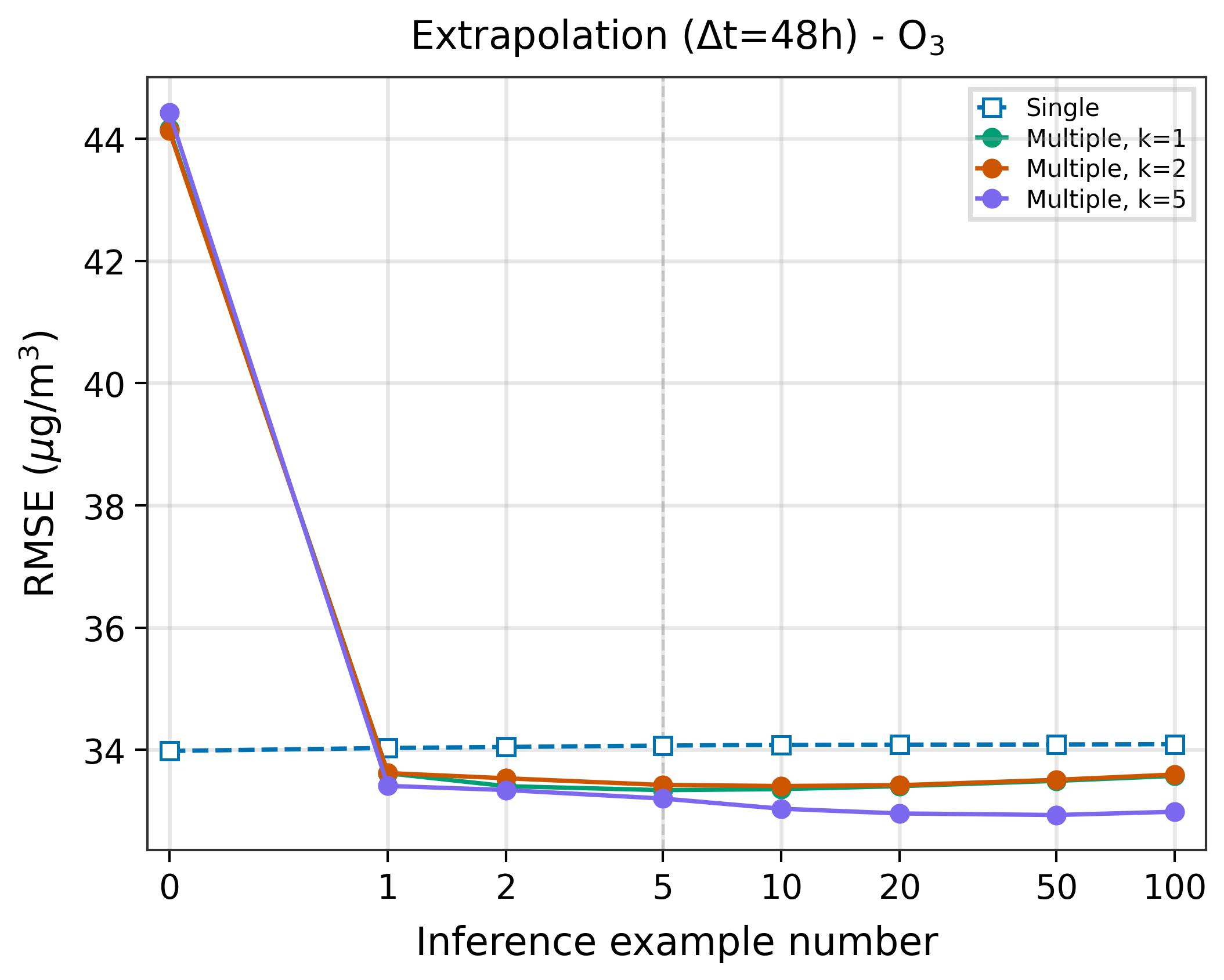}
    \end{minipage}
    \caption{Operator extrapolation to $\Delta t = 48$ (out-of-distribution) on BTHSA. Left: PM$_{2.5}$. Right: O$_3$. Single-operator shows flat curves, while example-trained ICON models improve with examples. Models with $k = 5$ achieve best extrapolation, with sustained improvement for PM$_{2.5}$ and a sharp initial drop for O$_3$.}
    \label{fig:extrapolation}
\end{figure}

Having established cardinality generalization for in-distribution operators, we now evaluate extrapolation to an unseen operator: models trained with $\Delta t \in [1, 24]$ are evaluated at $\Delta t = 48$ (out-of-distribution). We vary maximum example count $k \in \{1, 2, 5\}$ and evaluate performance as a function of validation example count (Figure~\ref{fig:extrapolation}). For comparison, we include the classical single-operator baseline ($\Delta t = 24$, trained without examples). It shows flat performance regardless of example count during inference.

ICON with operator diversity shows markedly different behavior. For O$_3$, all example-trained models exhibit a substantial RMSE drop with just one example. For PM$_{2.5}$, the initial drop is more modest, but error continues to decrease with additional examples. However, models trained with $k = 1$ or $2$ show slight performance degradation with larger example counts. In contrast, the model trained with $k = 5$ achieves the best extrapolation performance: RMSE generally decreases up to 100 examples with no signs of degradation.

When training spans multiple operators, each batch presents a different operator, so the model must rely on the provided examples to identify which operator is in play. This creates a natural incentive to extract operator-relevant information from examples rather than memorizing the training distribution, which could underlie both observations above. Ablation studies in Section~\ref{sec:ablation} further examine whether single-operator in-context learning can also benefit from examples within the in-context operator learning paradigm.

\subsection{Geometric Generalization}
\

\begin{figure}[t]
    \centering
    \begin{minipage}{0.48\textwidth}
        \centering
        \includegraphics[width=\linewidth]{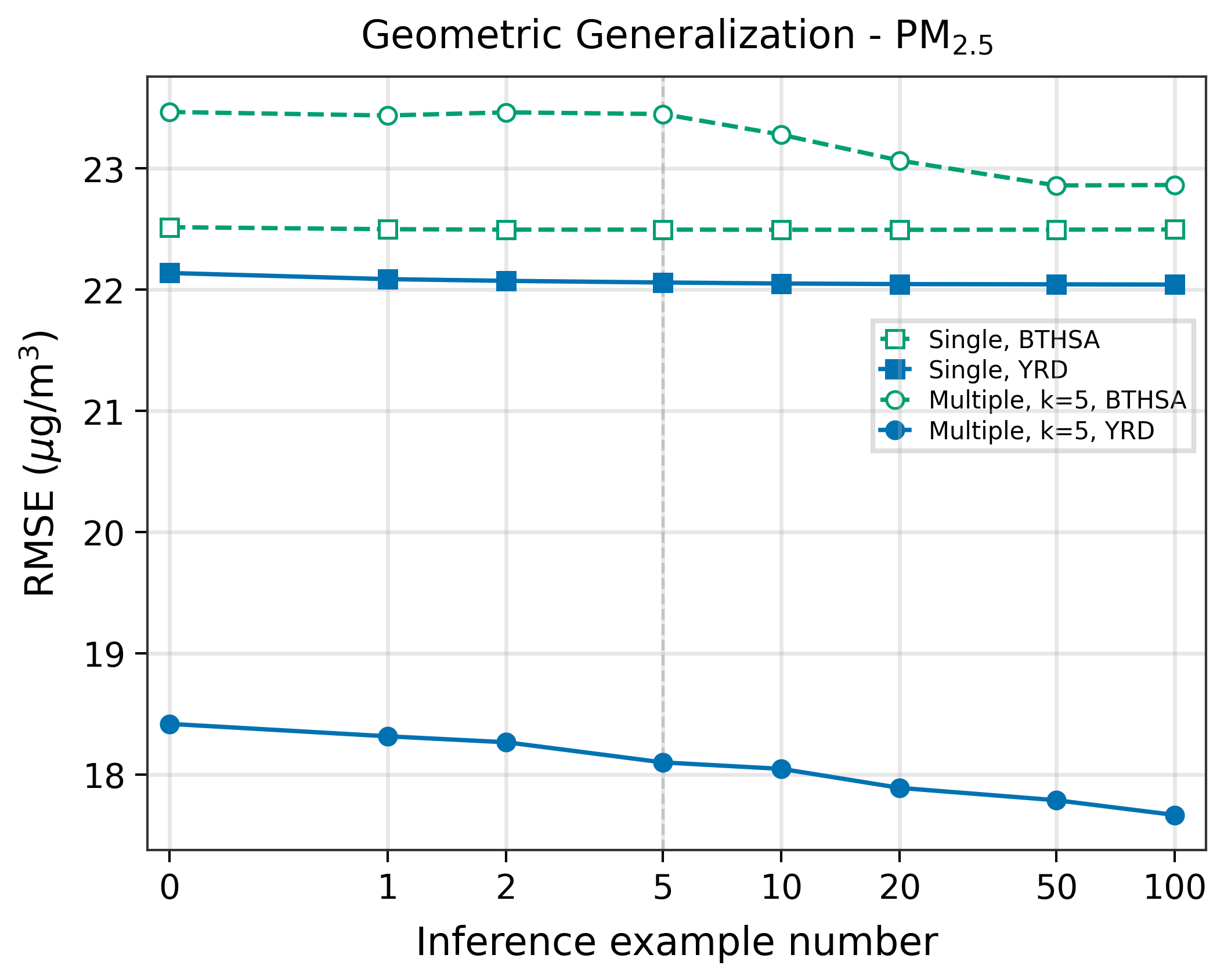}
    \end{minipage}
    \hfill
    \begin{minipage}{0.48\textwidth}
        \centering
        \includegraphics[width=\linewidth]{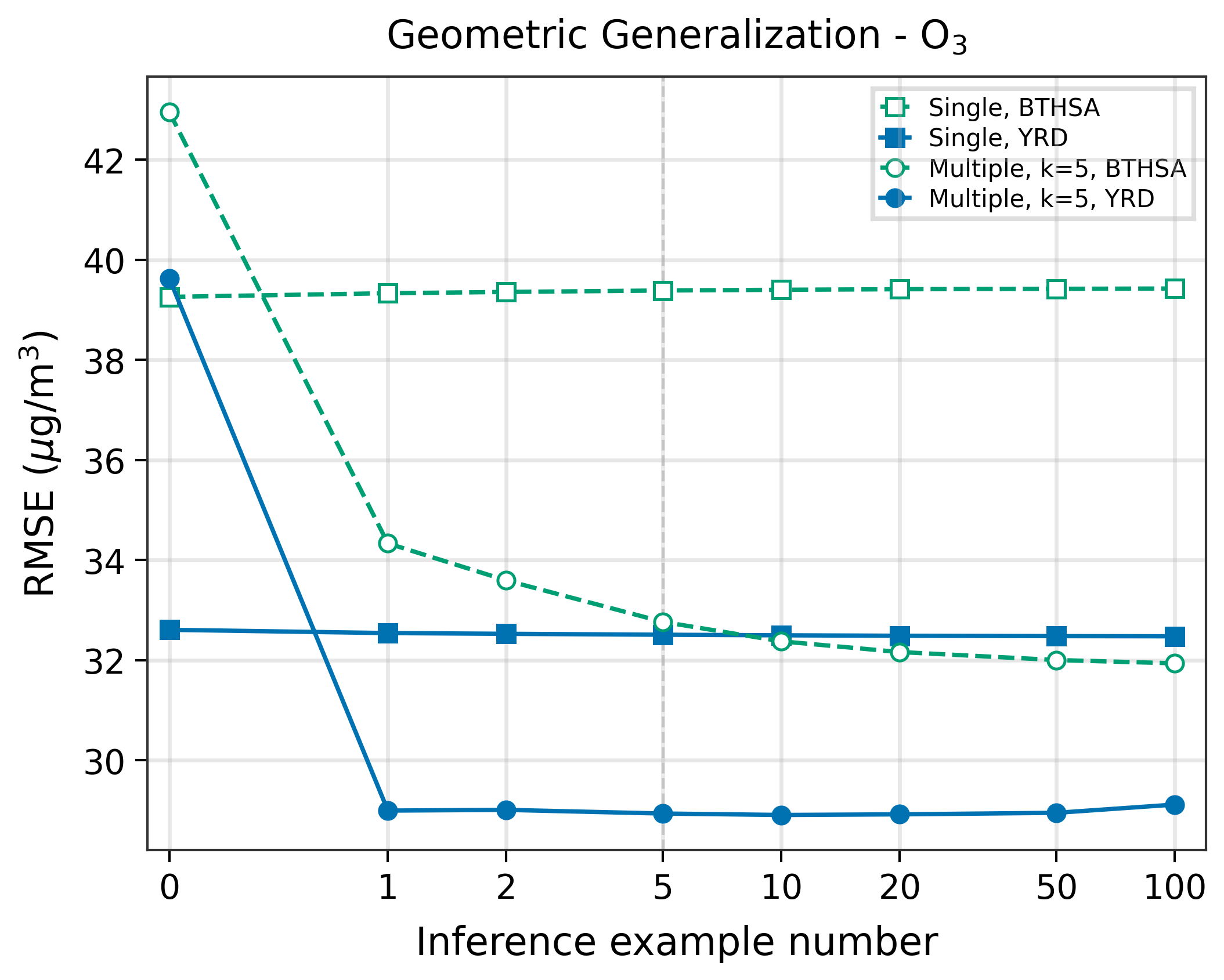}
    \end{minipage}
    \caption{Geometric generalization: models trained on BTHSA or YRD, both evaluated on YRD at $\Delta t = 24$h without fine-tuning. Left: PM$_{2.5}$. Right: O$_3$. BTHSA-trained models transfer to YRD with moderate performance gap for PM$_{2.5}$ and minimal gap for O$_3$, suggesting cross-region generalization.}
    \label{fig:geometric_gen}
\end{figure}

To further assess the transferability of ICON with operator diversity, we conduct cross-region experiments that test whether learned representations can transfer across different spatial domains. We train models separately on BTHSA (228 stations) and YRD (127 stations), then evaluate both on YRD at $\Delta t = 24$h without fine-tuning. This tests whether a model trained on one region can generalize to a different graph topology. Figure~\ref{fig:geometric_gen} shows the results.

For PM$_{2.5}$ prediction, BTHSA-trained ICON models transferred to YRD show a moderate performance gap compared to YRD-native ICON models, reflecting domain differences. Despite this gap, the transferred models maintain stable performance across example counts, suggesting that the learned operator representations can transfer across different graph topologies.

For O$_3$ prediction, the transfer gap is notably smaller. Both native and transferred ICON models show the characteristic substantial improvement with examples, while the classical single-operator baseline remains flat regardless of training region. Notably, the BTHSA-trained ICON model with sufficient examples surpasses the classical single-operator baseline trained natively on YRD, suggesting that multi-operator in-context learning with examples may compensate for domain mismatch. This reinforces the association between operator diversity and improved generalization across both operators and spatial domains.

\subsection{Ablation: Single-Operator Learning with Examples}
\label{sec:ablation}
\

The preceding sections show that ICON with operator diversity substantially outperforms classical single-operator learning on complex tasks. A natural question arises: can single-operator models also benefit from examples within the in-context operator learning paradigm? We investigate this through ablation studies that examine the effect of examples when operator diversity is absent. Training dynamics show that single-operator settings exhibit more overfitting than multi-operator training (Appendix~\ref{sec:appendix-training-dynamics}).

\begin{figure}[t]
    \centering
    \begin{minipage}{0.48\textwidth}
        \centering
        \includegraphics[width=\linewidth]{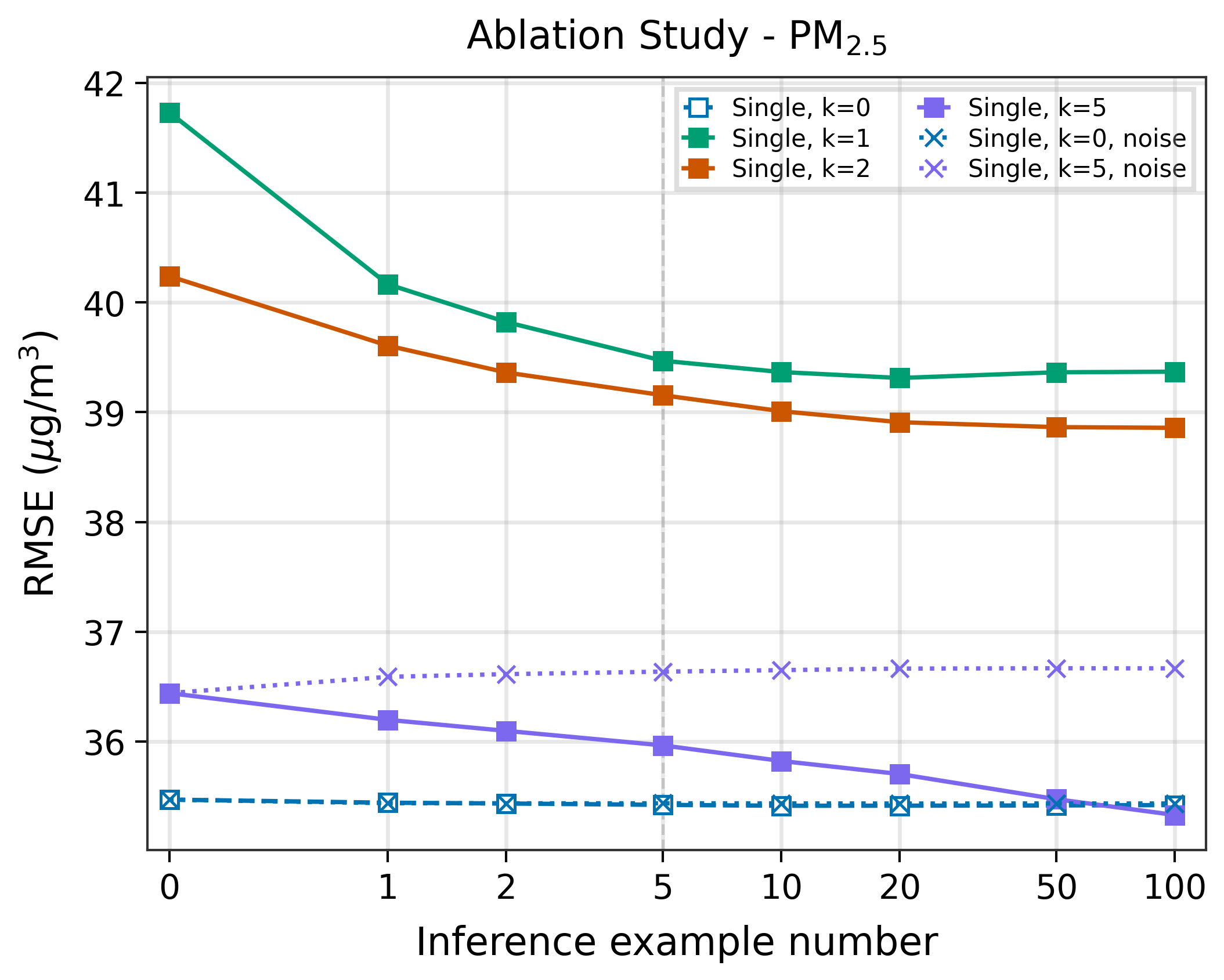}
    \end{minipage}
    \hfill
    \begin{minipage}{0.48\textwidth}
        \centering
        \includegraphics[width=\linewidth]{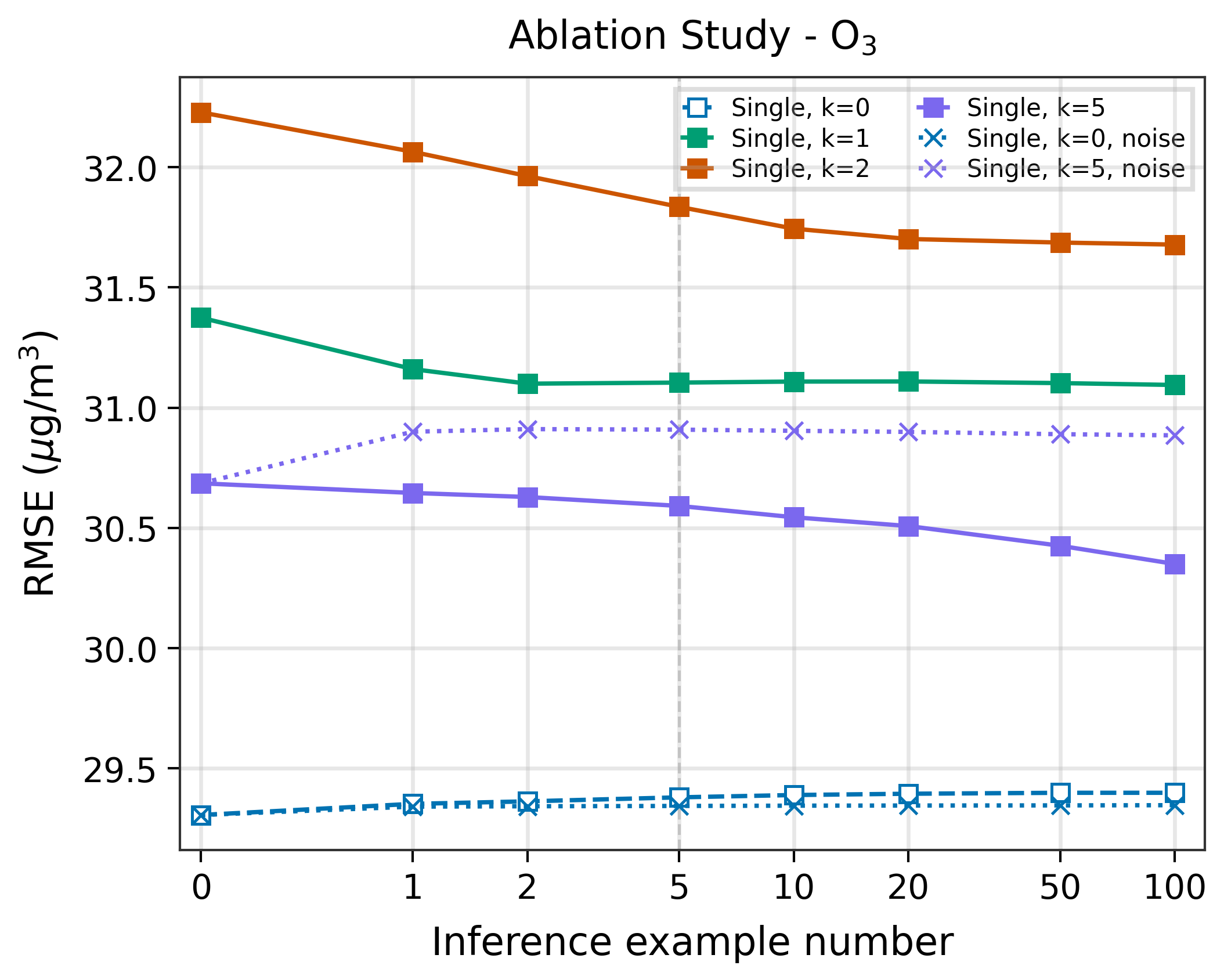}
    \end{minipage}
    \caption{Single-operator setting ($\Delta t = 24$, on BTHSA). Left: PM$_{2.5}$. Right: O$_3$. Models trained without contextual examples show no difference between noise and quality examples. The model trained with $k = 5$ approaches and slightly surpasses the baseline for PM$_{2.5}$ at 100 quality examples. However, when given noise, they show clear degradation, confirming that these models attend to example content.}
    \label{fig:single_op_noise}
\end{figure}

To complement the ICON cardinality results in Section~\ref{sec:cardinality}, we examine how models trained with single-operator examples scale with example count (Figure~\ref{fig:single_op_noise}). Providing examples during inference does improve performance over the zero-example baseline for all models trained with single-operator examples. In particular, the model trained with $k = 5$ exhibits continued improvement as example count increases and approaches the classical single-operator baseline, slightly surpassing it for PM$_{2.5}$ at 100 examples. To further probe whether examples provide meaningful information, we compare inference with high-quality examples versus random Gaussian noise. For the model trained without contextual examples ($k = 0$), noise examples perform comparably to high-quality ones, suggesting the model has learned to ignore example content entirely. In contrast, models trained with examples show performance degradation when given noise, indicating sensitivity to example content.

Together, the cardinality scaling and noise sensitivity results show some signal that single-operator models can learn to utilize examples, particularly when trained with sufficient example count ($k = 5$). However, the improvement is limited---the margin over the classical baseline is only observed for PM$_{2.5}$ and not for O$_3$---and compared to ICON with operator diversity, single-operator models require more examples to compensate for generally poorer performance, and are more prone to overfitting (Appendix~\ref{sec:appendix-training-dynamics}). How exactly examples are utilized within the model's forward pass, and how to further exploit in-context examples in single-operator learning scenarios, remain open questions for future investigation. These results nonetheless suggest that operator diversity provides a stronger learning signal for leveraging in-context examples under our current paradigm.

\section{Conclusion}
\

We presented a systematic comparison of in-context operator learning against classical single-operator learning, demonstrating the effectiveness of the in-context operator learning paradigm on real-world spatiotemporal systems. To enable this investigation, we introduced GICON (Graph In-Context Operator Network), which combines graph message passing for geometric generalization with example-aware positional encoding for cardinality generalization. These architectural innovations allow GICON to operate on irregularly sampled physical data and to scale to far more examples than seen during training---capabilities important for characterizing the scaling behavior of in-context learning. Our experiments on air quality prediction further demonstrate geometric generalization across regions with different graph topologies, and robust cardinality generalization from 0--5 training examples to 100 at inference, both of which support the advantage of in-context operator learning on complex tasks.

Our central finding is that in-context operator learning with operator diversity substantially outperforms classical single-operator learning on complex tasks: examples become increasingly informative with count, enabling general performance improvement even for out-of-distribution operators, while classical single-operator models show no such benefit. Our results indicate that operator diversity plays a crucial role in the in-context operator learning paradigm, substantially improving the utilization of in-context examples. Ablation studies show some signal that single-operator in-context learning can also benefit from examples, but the improvement is limited and single-operator models are more prone to overfitting. How examples are utilized within the model's forward pass, and how to further exploit in-context examples in single-operator scenarios, remain open questions. These results nonetheless suggest that operator diversity provides a stronger learning signal for leveraging in-context examples under our current paradigm.

Several directions merit further investigation. First, while we validated GICON on air quality prediction, the framework is applicable to a broader range of physical systems. Domains such as fluid dynamics, which exhibit rich multi-scale interactions and geometric symmetries, represent natural extensions where graph in-context operator learning may provide substantial advantages over grid-based methods. Second, extreme event prediction remains an important challenge in physical modeling; it remains practically important to investigate whether in-context operator learning can better capture rare, high-impact events. Third, our current example selection relies on similarity-based retrieval; developing more sophisticated strategies that account for operator-level diversity or uncertainty quantification could further improve the quality and efficiency of in-context learning. Finally, understanding why operator diversity so strongly facilitates example utilization, and developing strategies that enable single-operator in-context learning to leverage examples more robustly, remain promising directions for broadening the applicability of the in-context operator learning paradigm.

\section*{Acknowledgements}
Liu Yang acknowledges support from the National Research Foundation, Singapore, under the NRF fellowship (Project No. NRF-NRFF17-2025-0006). We acknowledge NUS IT's Research Computing group for providing computational support.

\bibliographystyle{unsrt}
\bibliography{references}

@inproceedings{zhang2019root,
  title     = {Root Mean Square Layer Normalization},
  author    = {Zhang, Biao and Sennrich, Rico},
  booktitle = {Advances in Neural Information Processing Systems (NeurIPS)},
  volume    = {32},
  year      = {2019}
}

@misc{jordan2024muon,
  title        = {Muon: An optimizer for hidden layers in neural networks},
  author       = {Jordan, Keller and Jin, Yuchen and Boza, Vlado and You, Jiacheng and Cesista, Franz and Newhouse, Laker and Bernstein, Jeremy},
  year         = {2024},
  howpublished = {\url{https://github.com/KellerJordan/Muon}}
}

@article{e2017deep,
  title     = {Deep learning-based numerical methods for high-dimensional parabolic partial differential equations and backward stochastic differential equations},
  author    = {E, Weinan and Han, Jiequn and Jentzen, Arnulf},
  journal   = {Communications in Mathematics and Statistics},
  volume    = {5},
  number    = {4},
  pages     = {349--380},
  year      = {2017},
  publisher = {Springer}
}

@article{han2018solving,
  title     = {Solving high-dimensional partial differential equations using deep learning},
  author    = {Han, Jiequn and Jentzen, Arnulf and E, Weinan},
  journal   = {Proceedings of the National Academy of Sciences},
  volume    = {115},
  number    = {34},
  pages     = {8505--8510},
  year      = {2018},
  publisher = {National Acad Sciences}
}

@article{raissi2019physics,
  title     = {Physics-informed neural networks: A deep learning framework for solving forward and inverse problems involving nonlinear partial differential equations},
  author    = {Raissi, Maziar and Perdikaris, Paris and Karniadakis, George Em},
  journal   = {Journal of Computational Physics},
  volume    = {378},
  pages     = {686--707},
  year      = {2019},
  publisher = {Elsevier}
}

@article{sirignano2018dgm,
  title     = {DGM: A deep learning algorithm for solving partial differential equations},
  author    = {Sirignano, Justin and Spiliopoulos, Konstantinos},
  journal   = {Journal of Computational Physics},
  volume    = {375},
  pages     = {1339--1364},
  year      = {2018},
  publisher = {Elsevier}
}

@article{weinan2018deep,
  title     = {The deep Ritz method: A deep learning-based numerical algorithm for solving variational problems},
  author    = {E, Weinan and Yu, Bing},
  journal   = {Communications in Mathematics and Statistics},
  volume    = {6},
  number    = {1},
  pages     = {1--12},
  year      = {2018},
  publisher = {Springer}
}

@article{cao2024vicon,
  title   = {{VICON}: Vision In-Context Operator Networks for Multi-Physics Fluid Dynamics Prediction},
  author  = {Cao, Yadi and Liu, Yuxuan and Yang, Liu and Yu, Rose and Schaeffer, Hayden and Osher, Stanley J.},
  journal = {Transactions on Machine Learning Research},
  year    = {2026},
  url     = {https://openreview.net/forum?id=6V3YmHULQ3}
}

@article{douze2025faiss,
  title     = {The {Faiss} Library},
  author    = {Douze, Matthijs and Guzhva, Alexandr and Deng, Chengqi and Johnson, Jeff and Szilvasy, Gergely and Mazar{\'e}, Pierre-Emmanuel and Lomeli, Maria and Hosseini, Lucas and J{\'e}gou, Herv{\'e}},
  journal   = {IEEE Transactions on Big Data},
  year      = {2025},
  doi       = {10.1109/TBDATA.2025.3618474},
  publisher = {IEEE}
}

@inproceedings{long2018pdenet,
  title     = {{PDE-Net}: Learning {PDEs} from Data},
  author    = {Long, Zichao and Lu, Yiping and Ma, Xianzhong and Dong, Bin},
  booktitle = {International Conference on Machine Learning (ICML)},
  pages     = {3208--3216},
  year      = {2018},
  publisher = {PMLR}
}

@article{chen1995approximation,
  title     = {Approximation capability to functions of several variables, nonlinear functionals, and operators by radial basis function neural networks},
  author    = {Chen, Tianping and Chen, Hong},
  journal   = {IEEE Transactions on Neural Networks},
  volume    = {6},
  number    = {4},
  pages     = {904--910},
  year      = {1995},
  publisher = {IEEE}
}

@article{chen1995universal,
  title     = {Universal approximation to nonlinear operators by neural networks with arbitrary activation functions and its application to dynamical systems},
  author    = {Chen, Tianping and Chen, Hong},
  journal   = {IEEE Transactions on Neural Networks},
  volume    = {6},
  number    = {4},
  pages     = {911--917},
  year      = {1995},
  publisher = {IEEE}
}

@inproceedings{li2021fourier,
  title     = {Fourier Neural Operator for Parametric Partial Differential Equations},
  author    = {Li, Zongyi and Kovachki, Nikola and Azizzadenesheli, Kamyar and Liu, Burigede and Bhattacharya, Kaushik and Stuart, Andrew and Anandkumar, Anima},
  booktitle = {International Conference on Learning Representations (ICLR)},
  year      = {2021},
  url       = {https://openreview.net/forum?id=c8P9NQVtmnO}
}

@article{kovachki2023neural,
  title     = {Neural operator: Learning maps between function spaces with applications to {PDE}s},
  author    = {Kovachki, Nikola and Li, Zongyi and Liu, Burigede and Azizzadenesheli, Kamyar and Bhattacharya, Kaushik and Stuart, Andrew and Anandkumar, Anima},
  journal   = {Journal of Machine Learning Research},
  volume    = {24},
  number    = {89},
  pages     = {1--97},
  year      = {2023}
}

@article{lu2021learning,
  title     = {Learning nonlinear operators via {DeepONet} based on the universal approximation theorem of operators},
  author    = {Lu, Lu and Jin, Pengzhan and Pang, Guofei and Zhang, Zhongqiang and Karniadakis, George Em},
  journal   = {Nature Machine Intelligence},
  volume    = {3},
  number    = {3},
  pages     = {218--229},
  year      = {2021},
  publisher = {Nature Publishing Group}
}

@article{lu2022comprehensive,
  title     = {A comprehensive and fair comparison of two neural operators (with practical extensions) based on fair data},
  author    = {Lu, Lu and Meng, Xuhui and Cai, Shengze and Mao, Zhiping and Goswami, Somdatta and Zhang, Zhongqiang and Karniadakis, George Em},
  journal   = {Computer Methods in Applied Mechanics and Engineering},
  volume    = {393},
  pages     = {114778},
  year      = {2022},
  publisher = {Elsevier}
}

@article{jin2022mionet,
  title     = {{MIONet}: Learning multiple-input operators via tensor product},
  author    = {Jin, Pengzhan and Meng, Shuai and Lu, Lu},
  journal   = {SIAM Journal on Scientific Computing},
  volume    = {44},
  number    = {6},
  pages     = {A3490--A3514},
  year      = {2022},
  publisher = {SIAM}
}

@article{wang2021learning,
  title     = {Learning the solution operator of parametric partial differential equations with physics-informed {DeepONet}s},
  author    = {Wang, Sifan and Wang, Hanwen and Perdikaris, Paris},
  journal   = {Science Advances},
  volume    = {7},
  number    = {40},
  pages     = {eabi8605},
  year      = {2021},
  publisher = {American Association for the Advancement of Science}
}

@article{zhang2024bayesian,
  title     = {Bayesian deep operator learning for homogenized to fine-scale maps for multiscale {PDE}},
  author    = {Zhang, Zecheng and Moya, Christian and Leung, Wing Tat and Lin, Guang and Schaeffer, Hayden},
  journal   = {Multiscale Modeling \& Simulation},
  volume    = {22},
  number    = {3},
  pages     = {956--972},
  year      = {2024},
  publisher = {SIAM}
}

@article{wen2022u,
  title     = {{U-FNO}---an enhanced Fourier neural operator-based deep-learning model for multiphase flow},
  author    = {Wen, Gege and Li, Zongyi and Azizzadenesheli, Kamyar and Anandkumar, Anima and Benson, Sally M},
  journal   = {Advances in Water Resources},
  volume    = {163},
  pages     = {104180},
  year      = {2022},
  publisher = {Elsevier}
}

@article{zhang2023belnet,
  title     = {{BelNet}: Basis enhanced learning, a mesh-free neural operator},
  author    = {Zhang, Zecheng and Leung, Wing Tat and Schaeffer, Hayden},
  journal   = {Proceedings of the Royal Society A},
  volume    = {479},
  number    = {2276},
  pages     = {20230043},
  year      = {2023},
  publisher = {The Royal Society}
}

@inproceedings{li2023geometry,
  title     = {Geometry-informed neural operator for large-scale {3D} {PDE}s},
  author    = {Li, Zongyi and Kovachki, Nikola and Choy, Chris and Li, Boyi and Kossaifi, Jean and Otta, Shourya and Nabian, Mohammad Amin and Stadler, Maximilian and Hundt, Christian and Azizzadenesheli, Kamyar and others},
  booktitle = {Advances in Neural Information Processing Systems (NeurIPS)},
  volume    = {36},
  year      = {2023}
}

@article{zhang2024d2no,
  title     = {{D2NO}: Efficient handling of heterogeneous input function spaces with distributed deep neural operators},
  author    = {Zhang, Zecheng and Moya, Christian and Lu, Lu and Lin, Guang and Schaeffer, Hayden},
  journal   = {Computer Methods in Applied Mechanics and Engineering},
  volume    = {428},
  pages     = {117084},
  year      = {2024},
  publisher = {Elsevier}
}

@inproceedings{kurth2023fourcastnet,
  title     = {{FourCastNet}: Accelerating Global High-Resolution Weather Forecasting Using Adaptive {Fourier} Neural Operators},
  author    = {Kurth, Thorsten and Subramanian, Shashank and Harrington, Peter and Pathak, Jaideep and Mardani, Morteza and Hall, David and Miele, Andrea and Kashinath, Karthik and Anandkumar, Anima},
  booktitle = {Proceedings of the Platform for Advanced Scientific Computing Conference (PASC '23)},
  year      = {2023},
  address   = {Davos, Switzerland},
  publisher = {ACM},
  doi       = {10.1145/3592979.3593412}
}

@article{jiang2024fourier,
  title     = {Fourier-{MIONet}: {F}ourier-enhanced multiple-input neural operators for multiphase modeling of geological carbon sequestration},
  author    = {Jiang, Zhongyi and Zhu, Min and Lu, Lu},
  journal   = {Reliability Engineering \& System Safety},
  volume    = {251},
  pages     = {110392},
  year      = {2024},
  publisher = {Elsevier}
}

@article{yin2024dimon,
  title     = {A scalable framework for learning the geometry-dependent solution operators of partial differential equations},
  author    = {Yin, Minglang and Charon, Nicolas and Brody, Ryan and Lu, Lu and Trayanova, Natalia A. and Maggioni, Mauro},
  journal   = {Nature Computational Science},
  volume    = {4},
  number    = {12},
  pages     = {928--940},
  year      = {2024},
  publisher = {Nature Publishing Group},
  doi       = {10.1038/s43588-024-00732-2}
}

@article{moya2025conformalized,
  title     = {Conformalized-{DeepONet}: A distribution-free framework for uncertainty quantification in deep operator networks},
  author    = {Moya, Christian and Mollaali, Amirhossein and Zhang, Zecheng and Lu, Lu and Lin, Guang},
  journal   = {Physica D: Nonlinear Phenomena},
  volume    = {471},
  pages     = {134418},
  year      = {2025},
  publisher = {Elsevier},
  doi       = {10.1016/j.physd.2024.134418}
}

@article{yang2023context,
  title     = {In-context operator learning with data prompts for differential equation problems},
  author    = {Yang, Liu and Liu, Siting and Meng, Tingwei and Osher, Stanley J.},
  journal   = {Proceedings of the National Academy of Sciences},
  volume    = {120},
  number    = {39},
  pages     = {e2310142120},
  year      = {2023},
  publisher = {National Academy of Sciences}
}

@article{yang2025fine,
  title     = {Fine-tune language models as multi-modal differential equation solvers},
  author    = {Yang, Liu and Liu, Siting and Osher, Stanley J.},
  journal   = {Neural Networks},
  year      = {2025},
  publisher = {Elsevier},
  doi       = {10.1016/j.neunet.2025.107455}
}

@article{yang2024pde,
  title     = {{PDE} generalization of in-context operator networks: A study on {1D} scalar nonlinear conservation laws},
  author    = {Yang, Liu and Osher, Stanley J.},
  journal   = {Journal of Computational Physics},
  volume    = {519},
  pages     = {113379},
  year      = {2024},
  publisher = {Elsevier},
  doi       = {10.1016/j.jcp.2024.113379}
}

@inproceedings{liu2023does,
  title     = {Does In-Context Operator Learning Generalize to Domain-Shifted Settings?},
  author    = {Liu, Jerry Weihong and Erichson, N. Benjamin and Bhatia, Kush and Mahoney, Michael W. and Re, Christopher},
  booktitle = {NeurIPS 2023 Workshop on the Symbiosis of Deep Learning and Differential Equations (DLDE III)},
  year      = {2023}
}

@misc{zhang2025probabilistic,
  title         = {Probabilistic operator learning: generative modeling and uncertainty quantification for foundation models of differential equations},
  author        = {Zhang, Benjamin J. and Liu, Siting and Osher, Stanley J. and Katsoulakis, Markos A.},
  year          = {2025},
  eprint        = {2509.05186},
  archivePrefix = {arXiv},
  primaryClass  = {stat.ML},
  url           = {https://arxiv.org/abs/2509.05186}
}

@misc{meng2025solving,
  title         = {Solving Optimal Execution Problems via In-Context Operator Networks},
  author        = {Meng, Tingwei and Vo{\ss}, Moritz and Detering, Nils and Farolfi, Giulio and Osher, Stanley J. and Menz, Georg},
  year          = {2025},
  eprint        = {2501.15106},
  archivePrefix = {arXiv},
  primaryClass  = {q-fin.TR},
  url           = {https://arxiv.org/abs/2501.15106}
}

@misc{cole2024incontext,
  title         = {In-Context Learning of Linear Systems: Generalization Theory and Applications to Operator Learning},
  author        = {Cole, Frank and Lu, Yulong and Xu, Wuzhe and Zhang, Tianhao},
  year          = {2024},
  eprint        = {2409.12293},
  archivePrefix = {arXiv},
  primaryClass  = {cs.LG},
  url           = {https://arxiv.org/abs/2409.12293}
}

@misc{cole2026incontext,
  title         = {In-Context Operator Learning on the Space of Probability Measures},
  author        = {Cole, Frank and Wang, Dixi and Chen, Yineng and Lu, Yulong and Lai, Rongjie},
  year          = {2026},
  eprint        = {2601.09979},
  archivePrefix = {arXiv},
  primaryClass  = {cs.LG},
  url           = {https://arxiv.org/abs/2601.09979}
}

@misc{mishra2025continuum,
  title         = {Continuum Transformers Perform In-Context Learning by Operator Gradient Descent},
  author        = {Mishra, Abhiti and Patel, Yash and Tewari, Ambuj},
  year          = {2025},
  eprint        = {2505.17838},
  archivePrefix = {arXiv},
  primaryClass  = {stat.ML},
  url           = {https://arxiv.org/abs/2505.17838}
}

@misc{li2020neural,
  title         = {Neural Operator: Graph Kernel Network for Partial Differential Equations},
  author        = {Li, Zongyi and Kovachki, Nikola and Azizzadenesheli, Kamyar and Liu, Burigede and Bhattacharya, Kaushik and Stuart, Andrew and Anandkumar, Anima},
  year          = {2020},
  eprint        = {2003.03485},
  archivePrefix = {arXiv},
  primaryClass  = {cs.LG},
  url           = {https://arxiv.org/abs/2003.03485}
}

@inproceedings{li2020multipole,
  title     = {Multipole Graph Neural Operator for Parametric Partial Differential Equations},
  author    = {Li, Zongyi and Kovachki, Nikola and Azizzadenesheli, Kamyar and Liu, Burigede and Bhattacharya, Kaushik and Stuart, Andrew and Anandkumar, Anima},
  booktitle = {Advances in Neural Information Processing Systems (NeurIPS)},
  volume    = {33},
  pages     = {6755--6766},
  year      = {2020}
}

@inproceedings{mousavi2025rigno,
  title     = {{RIGNO}: A Graph-based Framework for Robust and Accurate Operator Learning for {PDE}s on Arbitrary Domains},
  author    = {Mousavi, Sepehr and Wen, Shizheng and Lingsch, Levi and Herde, Maximilian and Raoni{\'c}, Bogdan and Mishra, Siddhartha},
  booktitle = {Advances in Neural Information Processing Systems (NeurIPS)},
  volume    = {38},
  year      = {2025}
}

@misc{battaglia2018relationalinductivebiasesdeep,
  title         = {Relational inductive biases, deep learning, and graph networks},
  author        = {Battaglia, Peter W. and Hamrick, Jessica B. and Bapst, Victor and Sanchez-Gonzalez, Alvaro and Zambaldi, Vinicius and Malinowski, Mateusz and Tacchetti, Andrea and Raposo, David and Santoro, Adam and Faulkner, Ryan and Gulcehre, Caglar and Song, Francis and Ballard, Andrew and Gilmer, Justin and Dahl, George and Vaswani, Ashish and Allen, Kelsey and Nash, Charles and Langston, Victoria and Dyer, Chris and Heess, Nicolas and Wierstra, Daan and Kohli, Pushmeet and Botvinick, Matt and Vinyals, Oriol and Li, Yujia and Pascanu, Razvan},
  year          = {2018},
  eprint        = {1806.01261},
  archivePrefix = {arXiv},
  primaryClass  = {cs.LG},
  url           = {https://arxiv.org/abs/1806.01261}
}

@inproceedings{sanchezgonzalez2020learningsimulatecomplexphysics,
  title     = {Learning to Simulate Complex Physics with Graph Networks},
  author    = {Sanchez-Gonzalez, Alvaro and Godwin, Jonathan and Pfaff, Tobias and Ying, Rex and Leskovec, Jure and Battaglia, Peter W.},
  booktitle = {International Conference on Machine Learning (ICML)},
  volume    = {119},
  pages     = {8459--8468},
  year      = {2020},
  publisher = {PMLR}
}

@inproceedings{pfaff2021learningmeshbasedsimulationgraph,
  title     = {Learning Mesh-Based Simulation with Graph Networks},
  author    = {Pfaff, Tobias and Fortunato, Meire and Sanchez-Gonzalez, Alvaro and Battaglia, Peter W.},
  booktitle = {International Conference on Learning Representations (ICLR)},
  year      = {2021},
  url       = {https://openreview.net/forum?id=roNqYL0_XP},
  note      = {Outstanding Paper Award}
}

@inproceedings{brandstetter2022messagepassingneuralpde,
  title     = {Message Passing Neural {PDE} Solvers},
  author    = {Brandstetter, Johannes and Worrall, Daniel and Welling, Max},
  booktitle = {International Conference on Learning Representations (ICLR)},
  year      = {2022},
  url       = {https://openreview.net/forum?id=vSix3HPYKSU},
  note      = {Spotlight}
}

@misc{wang2025pcdcnet,
  title         = {{PCDCNet}: A Surrogate Model for Air Quality Forecasting with Physical-Chemical Dynamics and Constraints},
  author        = {Wang, Shuo and Cheng, Yun and Meng, Qingye and Saukh, Olga and Zhang, Jiang and Fan, Jingfang and Zhang, Yuanting and Yuan, Xingyuan and Thiele, Lothar},
  year          = {2025},
  eprint        = {2505.19842},
  archivePrefix = {arXiv},
  primaryClass  = {cs.LG},
  url           = {https://arxiv.org/abs/2505.19842}
}

@article{metpy,
  author    = {May, Ryan M. and Goebbert, Kevin H. and Thielen, Jonathan E. and Leeman, John R. and Camron, M. Drew and Bruick, Zachary and Bruning, Eric C. and Manser, Russell P. and Arms, Sean C. and Marsh, Patrick T.},
  title     = {{MetPy}: A Meteorological Python Library for Data Analysis and Visualization},
  journal   = {Bulletin of the American Meteorological Society},
  year      = {2022},
  volume    = {103},
  number    = {10},
  pages     = {E2273--E2284},
  publisher = {American Meteorological Society},
  doi       = {10.1175/BAMS-D-21-0125.1},
  url       = {https://journals.ametsoc.org/view/journals/bams/103/10/BAMS-D-21-0125.1.xml}
}

@misc{wang2025knowairv2,
  title     = {{KnowAir-V2}: A Benchmark Dataset for Air Quality Forecasting with {PCDCNet}},
  author    = {Wang, Shuo and Cheng, Yun and Meng, Qingye and Saukh, Olga and Zhang, Jiang and Fan, Jingfang and Zhang, Yuanting and Yuan, Xingyuan and Thiele, Lothar},
  year      = {2025},
  publisher = {Zenodo},
  doi       = {10.5281/zenodo.15614907},
  url       = {https://doi.org/10.5281/zenodo.15614907},
  note      = {Data set}
}

@techreport{radford2019language,
  title       = {Language Models are Unsupervised Multitask Learners},
  author      = {Radford, Alec and Wu, Jeffrey and Child, Rewon and Luan, David and Amodei, Dario and Sutskever, Ilya},
  institution = {OpenAI},
  year        = {2019}
}

@inproceedings{brown2020language,
  title     = {Language Models are Few-Shot Learners},
  author    = {Brown, Tom and Mann, Benjamin and Ryder, Nick and Subbiah, Melanie and Kaplan, Jared D and Dhariwal, Prafulla and Neelakantan, Arvind and Shyam, Pranav and Sastry, Girish and Askell, Amanda and others},
  booktitle = {Advances in Neural Information Processing Systems (NeurIPS)},
  volume    = {33},
  pages     = {1877--1901},
  year      = {2020}
}

@inproceedings{hao2023gnot,
  title     = {{GNOT}: A General Neural Operator Transformer for Operator Learning},
  author    = {Hao, Zhongkai and Wang, Zhengyi and Su, Hang and Ying, Chengyang and Dong, Yinpeng and Liu, Songming and Cheng, Ze and Song, Jian and Zhu, Jun},
  booktitle = {International Conference on Machine Learning (ICML)},
  volume    = {202},
  pages     = {12556--12569},
  year      = {2023},
  publisher = {PMLR}
}

@inproceedings{herde2024poseidon,
  title     = {Poseidon: Efficient Foundation Models for {PDE}s},
  author    = {Herde, Maximilian and Raoni{\'c}, Bogdan and Rohner, Tobias and K{\"a}ppeli, Roger and Molinaro, Roberto and de B{\'e}zenac, Emmanuel and Mishra, Siddhartha},
  booktitle = {Advances in Neural Information Processing Systems (NeurIPS)},
  volume    = {37},
  year      = {2024}
}

@misc{sun2024lemon,
  title         = {{LeMON}: Learning to Learn Multi-Operator Networks},
  author        = {Sun, Jingmin and Zhang, Zecheng and Schaeffer, Hayden},
  year          = {2024},
  eprint        = {2408.16168},
  archivePrefix = {arXiv},
  primaryClass  = {cs.LG},
  url           = {https://arxiv.org/abs/2408.16168}
}

@article{liu2024prose,
  title     = {PROSE: Predicting Operators and Symbolic Expressions using Multimodal Transformers},
  author    = {Liu, Yuxuan and Zhang, Zecheng and Schaeffer, Hayden},
  journal   = {Neural Networks},
  volume    = {180},
  pages     = {106707},
  year      = {2024},
  publisher = {Elsevier}
}

@inproceedings{liu2024proseFD,
  title     = {{PROSE-FD}: A Multimodal {PDE} Foundation Model for Learning Multiple Operators for Forecasting Fluid Dynamics},
  author    = {Liu, Yuxuan and Sun, Jingmin and He, Xinjie and Pinney, Griffin and Zhang, Zecheng and Schaeffer, Hayden},
  booktitle = {NeurIPS 2024 Workshop on Foundation Models for Science (FM4Science)},
  year      = {2024}
}

@inproceedings{serrano2025zebra,
  title     = {Zebra: In-Context and Generative Pretraining for Solving Parametric {PDEs}},
  author    = {Serrano, Louis and Koupa{\"i}, Armand Kassa{\"i} and Wang, Thomas X and Erbacher, Pierre and Gallinari, Patrick},
  booktitle = {International Conference on Machine Learning (ICML)},
  publisher = {PMLR},
  year      = {2025}
}

@inproceedings{chen2024dataefficient,
  title     = {Data-Efficient Operator Learning via Unsupervised Pretraining and In-Context Learning},
  author    = {Chen, Wuyang and Song, Jialin and Ren, Pu and Subramanian, Shashank and Morozov, Dmitriy and Mahoney, Michael},
  booktitle = {Advances in Neural Information Processing Systems (NeurIPS)},
  volume    = {37},
  year      = {2024}
}

@inproceedings{koupai2025enma,
  title     = {{ENMA}: Tokenwise Autoregression for Continuous Neural {PDE} Operators},
  author    = {Koupa{\"i}, Armand Kassa{\"i} and Le Boudec, Lise and Serrano, Louis and Gallinari, Patrick},
  booktitle = {Advances in Neural Information Processing Systems (NeurIPS)},
  volume    = {38},
  year      = {2025},
  url       = {https://openreview.net/forum?id=3CYXSMFv55}
}

\appendix

\clearpage
\section{Positional Encoding Algorithms}
\label{sec:appendix-pos-encoding}

\begin{algorithm}[H]
\caption{Positional Encoding for GICON}
\label{alg:pos-encoding}
\begin{algorithmic}
\Require Hidden states $\mathbf{H} \in \mathbb{R}^{|\mathcal{V}| \times (2k+1) \times d}$, number of attention heads $H$
\Statex
\Statex \textbf{--- Inter-Example Distinction: Example-Aware Attention Bias ---}
\Statex \Comment{Returns bias $\mathbf{A} \in \mathbb{R}^{H \times (2k+1) \times (2k+1)}$}
\State Extract key tokens at even positions: $\mathbf{H}_\mathbf{k} \gets \mathbf{H}[:, 0::2, :] \in \mathbb{R}^{|\mathcal{V}| \times (k+1) \times d}$
\State Pool across nodes: $\bar{\mathbf{H}}_\mathbf{k} \gets \text{mean}(\mathbf{H}_\mathbf{k}, \text{dim}=\text{nodes}) \in \mathbb{R}^{(k+1) \times d}$
\State Project to example embeddings: $\mathbf{Z} \gets \text{MLP}(\bar{\mathbf{H}}_\mathbf{k}) \in \mathbb{R}^{(k+1) \times d}$
\State Compute head-specific query/key: $\mathbf{Q}_z \gets \mathbf{W}_q\mathbf{Z}$, $\mathbf{K}_z \gets \mathbf{W}_k\mathbf{Z}$ \Comment{$\mathbf{W}_q, \mathbf{W}_k \in \mathbb{R}^{d \times d}$}
\State Compute pairwise similarities: $\mathbf{S} \gets \mathbf{Q}_z \mathbf{K}_z^\top / \sqrt{d/H} \in \mathbb{R}^{H \times (k+1) \times (k+1)}$
\State Initialize bias: $\mathbf{A} \gets \mathbf{0}^{H \times (2k+1) \times (2k+1)}$
\For{each position pair $(i, j)$ where $\lfloor i/2 \rfloor \leq k$ and $\lfloor j/2 \rfloor \leq k$}
    \State $\mathbf{A}[:, i, j] \gets \mathbf{S}[:, \lfloor i/2 \rfloor, \lfloor j/2 \rfloor]$ \Comment{Key-value pairs share same bias}
\EndFor
\Statex
\Statex \textbf{--- Key-Value Distinction: Input Mode ---}
\Statex \Comment{Learnable KV vector $\mathbf{r} \in \mathbb{R}^d$}
\State Create KV masks: $\mathbf{M}_\mathbf{k} \gets (t \bmod 2 = 0)$, $\mathbf{M}_\mathbf{v} \gets (t \bmod 2 = 1)$ for $t \in [0, 2k]$
\State Apply symmetric offsets: $\tilde{\mathbf{H}} \gets \mathbf{H} + \mathbf{M}_\mathbf{k} \cdot \mathbf{r} - \mathbf{M}_\mathbf{v} \cdot \mathbf{r}$
\State \Return $\tilde{\mathbf{H}}$ \Comment{Keys shifted by $+\mathbf{r}$, values by $-\mathbf{r}$}
\end{algorithmic}
\end{algorithm}

\clearpage
\section{Training Setup}
\label{sec:appendix-training-config}

\begin{table}[H]
\centering
\begin{minipage}{0.48\textwidth}
\centering
\caption{GICON model configuration.}
\label{tab:gicon_config}
\begin{tabular}{ll}
\toprule
\textbf{Component} & \textbf{Configuration} \\
\midrule
\multicolumn{2}{l}{\textit{Graph In-Context Operator Network}} \\
GICON layers & 3 \\
\midrule
\multicolumn{2}{l}{\textit{Message Passing Neural Network}} \\
Node dimension & 128 \\
Edge dimension & 128 \\
Message dimension & 256 \\
\midrule
\multicolumn{2}{l}{\textit{In-Context Transformer}} \\
Attention heads & 4 \\
Feed-forward dimension & 512 \\
Dropout & 0.1 \\
Normalization & RMSNorm~\cite{zhang2019root} \\
\bottomrule
\end{tabular}
\end{minipage}
\hfill
\begin{minipage}{0.48\textwidth}
\centering
\caption{Optimizer and learning rate schedule.}
\label{tab:optimizer_config}
\begin{tabular}{ll}
\toprule
\textbf{Parameter} & \textbf{Value} \\
\midrule
\multicolumn{2}{l}{\textit{Muon Optimizer~\cite{jordan2024muon}}} \\
Learning rate & $1 \times 10^{-4}$ \\
Weight decay & 0.01 \\
Gradient clipping & 1.0 \\
\midrule
\multicolumn{2}{l}{\textit{Cosine Decay Schedule}} \\
Warmup & 10\% of training \\
Decay period & 100\% of training \\
End learning rate factor & 0.1 \\
\bottomrule
\end{tabular}
\end{minipage}
\end{table}

Table~\ref{tab:gicon_config} summarizes the GICON architecture, and Table~\ref{tab:optimizer_config} details the optimization setup.

\medskip
\textbf{Loss Function.} The training objective is the mean squared error (MSE) computed over predicted pollution concentrations at all key positions and the query position in the sequence. Given $k$ in-context examples and one query, the causal attention mask enables the model to produce predictions $\tilde{\mathbf{v}}^{(1)}, \tilde{\mathbf{v}}^{(2)}, \ldots, \tilde{\mathbf{v}}^{(k)}$ at example key positions and $\tilde{\mathbf{v}}_{\text{query}}$ at the query position in a single forward pass. We extract the pollution channels (the last two channels) from each predicted value as $\tilde{\mathbf{p}}^{(j)}$ and similarly $\mathbf{p}^{(j)}$ from the ground truth $\mathbf{v}^{(j)}$. The loss is
\begin{equation}
\mathcal{L} = \text{MSE}\ (\tilde{\mathbf{p}},\mathbf{p}),
\end{equation}
where $\tilde{\mathbf{p}}=[\tilde{\mathbf{p}}^{(1)}; \tilde{\mathbf{p}}^{(2)}; \ldots, \tilde{\mathbf{p}}^{(k)}; \tilde{\mathbf{p}}_{\text{query}}]$ and $\mathbf{p}=[\mathbf{p}^{(1)}; \mathbf{p}^{(2)}; \ldots; \mathbf{p}^{(k)}; \mathbf{p}_{\text{label}}]$ denote the concatenation of predicted and ground-truth pollution values across all $k$ in-context examples and the query.

\clearpage
\section{Training Dynamics}
\label{sec:appendix-training-dynamics}

\begin{figure}[H]
    \centering
    \begin{minipage}{0.48\textwidth}
        \centering
        \includegraphics[width=\linewidth]{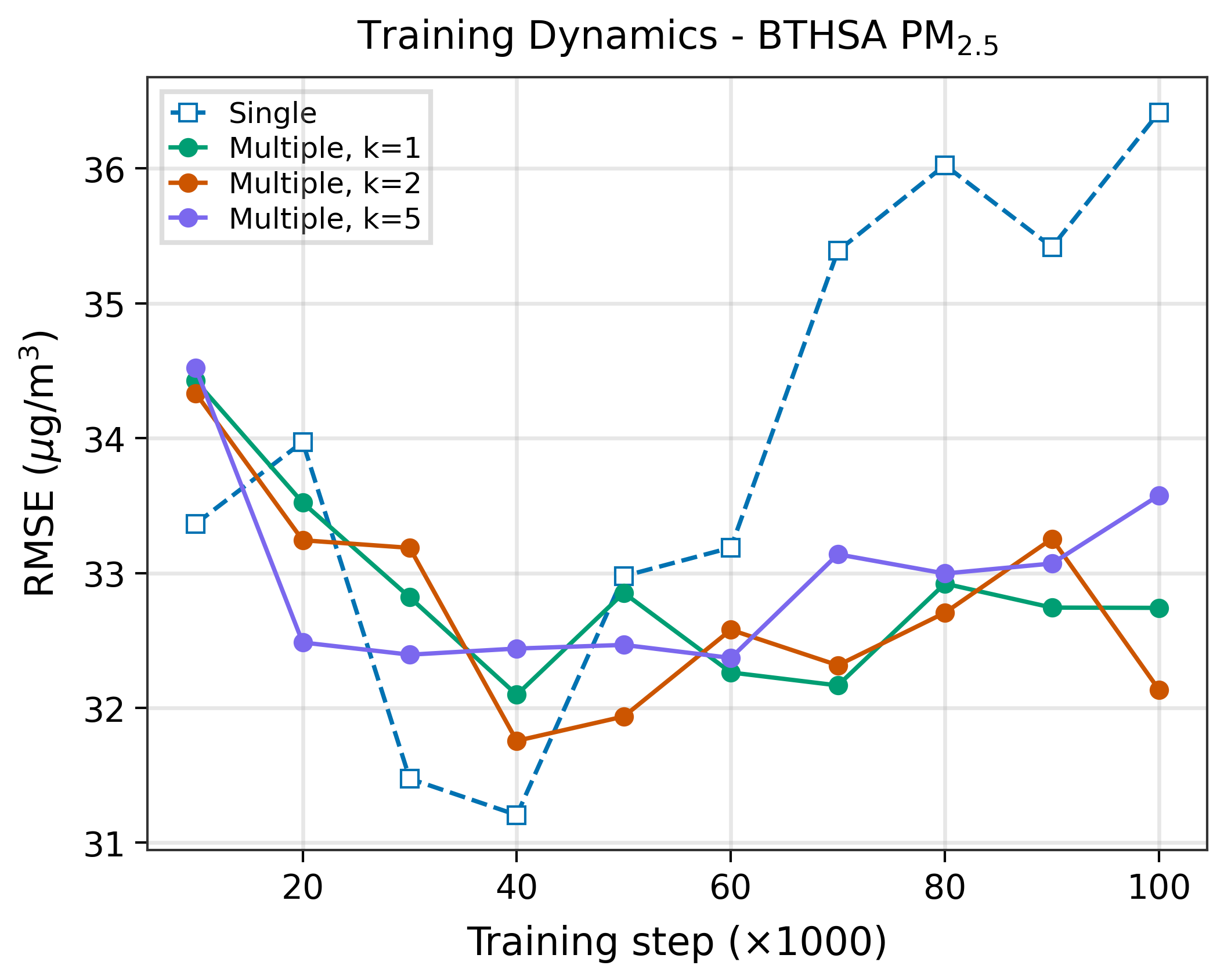}
    \end{minipage}
    \hfill
    \begin{minipage}{0.48\textwidth}
        \centering
        \includegraphics[width=\linewidth]{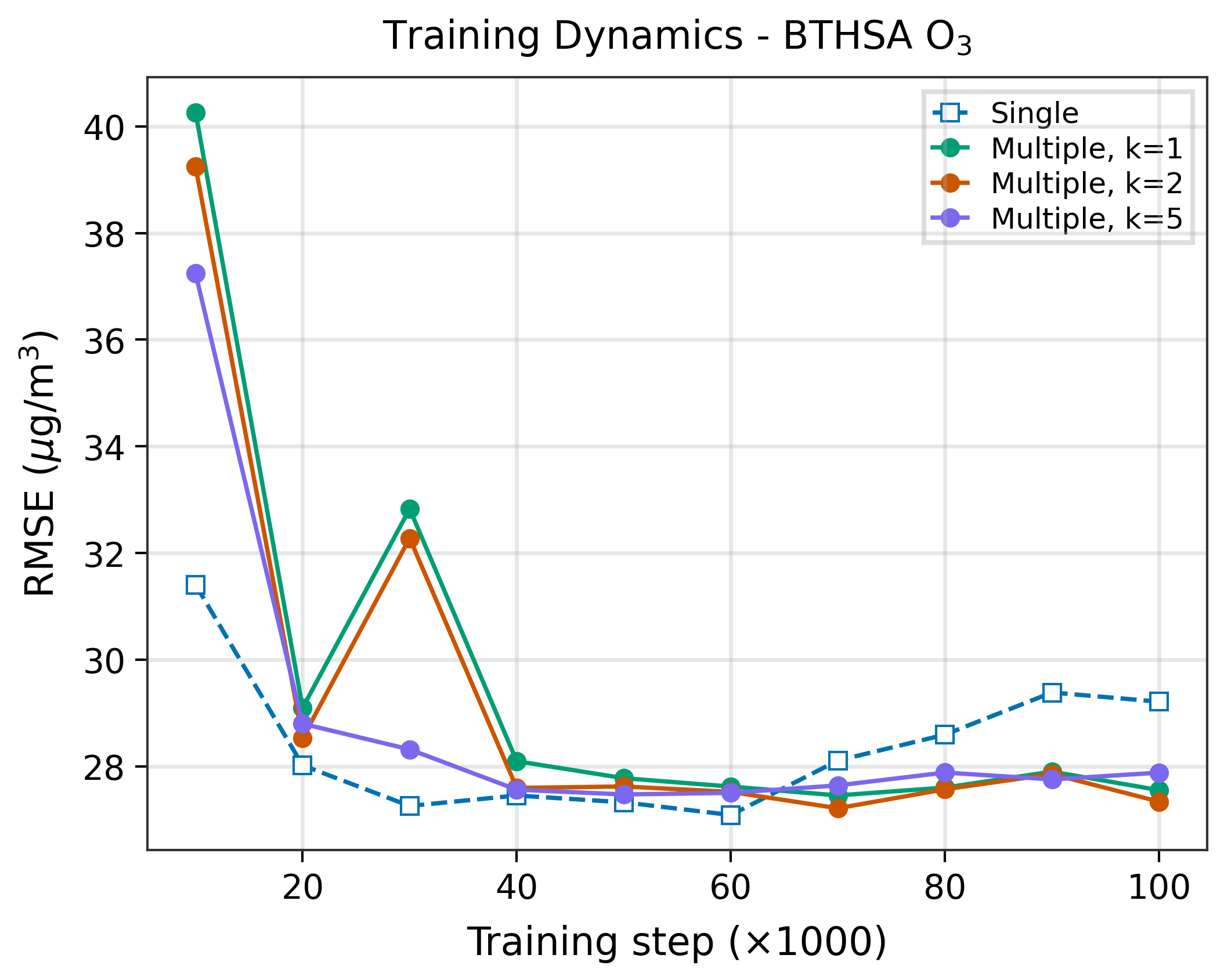}
    \end{minipage}
    \\[0.3em]
    \begin{minipage}{0.48\textwidth}
        \centering
        \includegraphics[width=\linewidth]{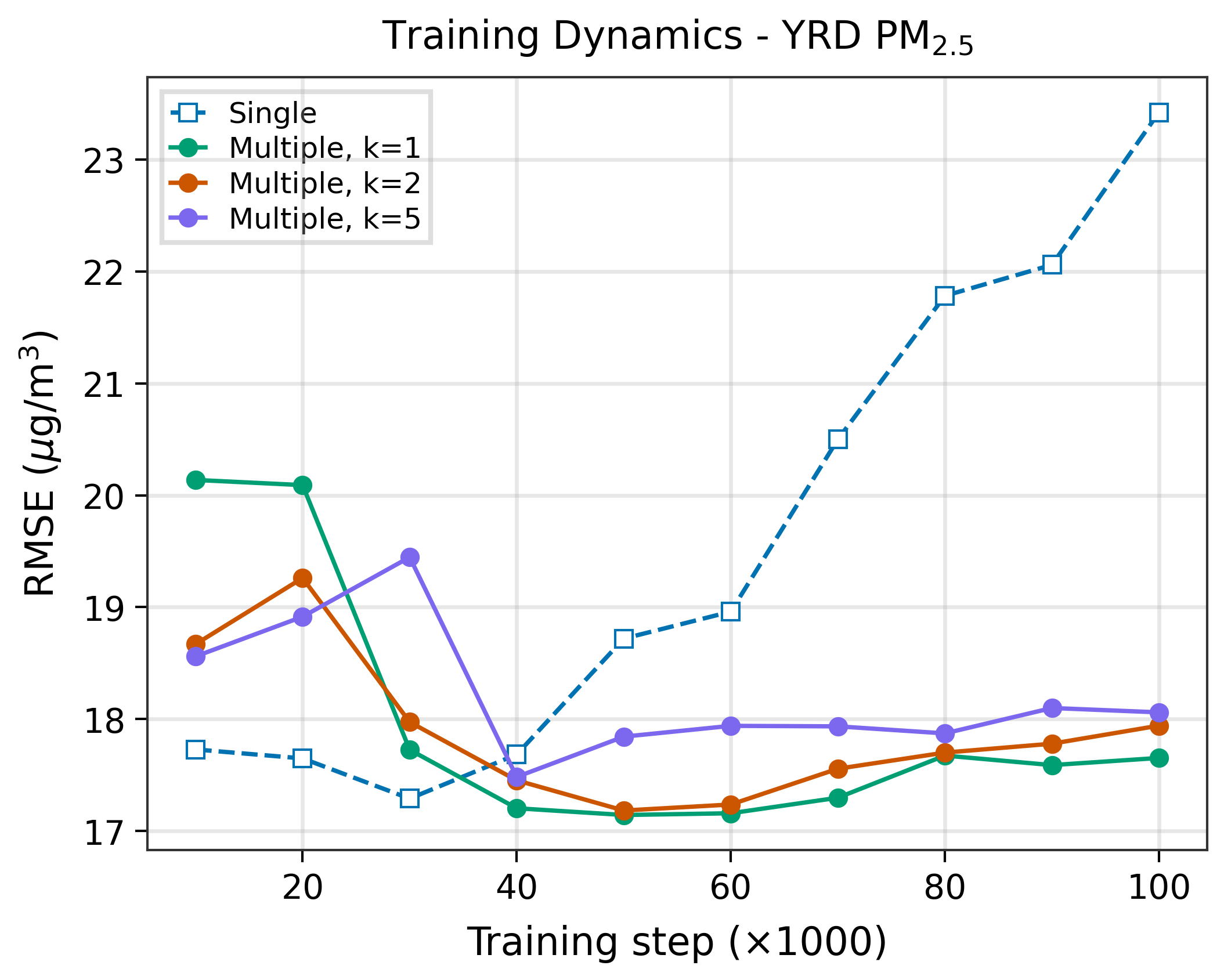}
    \end{minipage}
    \hfill
    \begin{minipage}{0.48\textwidth}
        \centering
        \includegraphics[width=\linewidth]{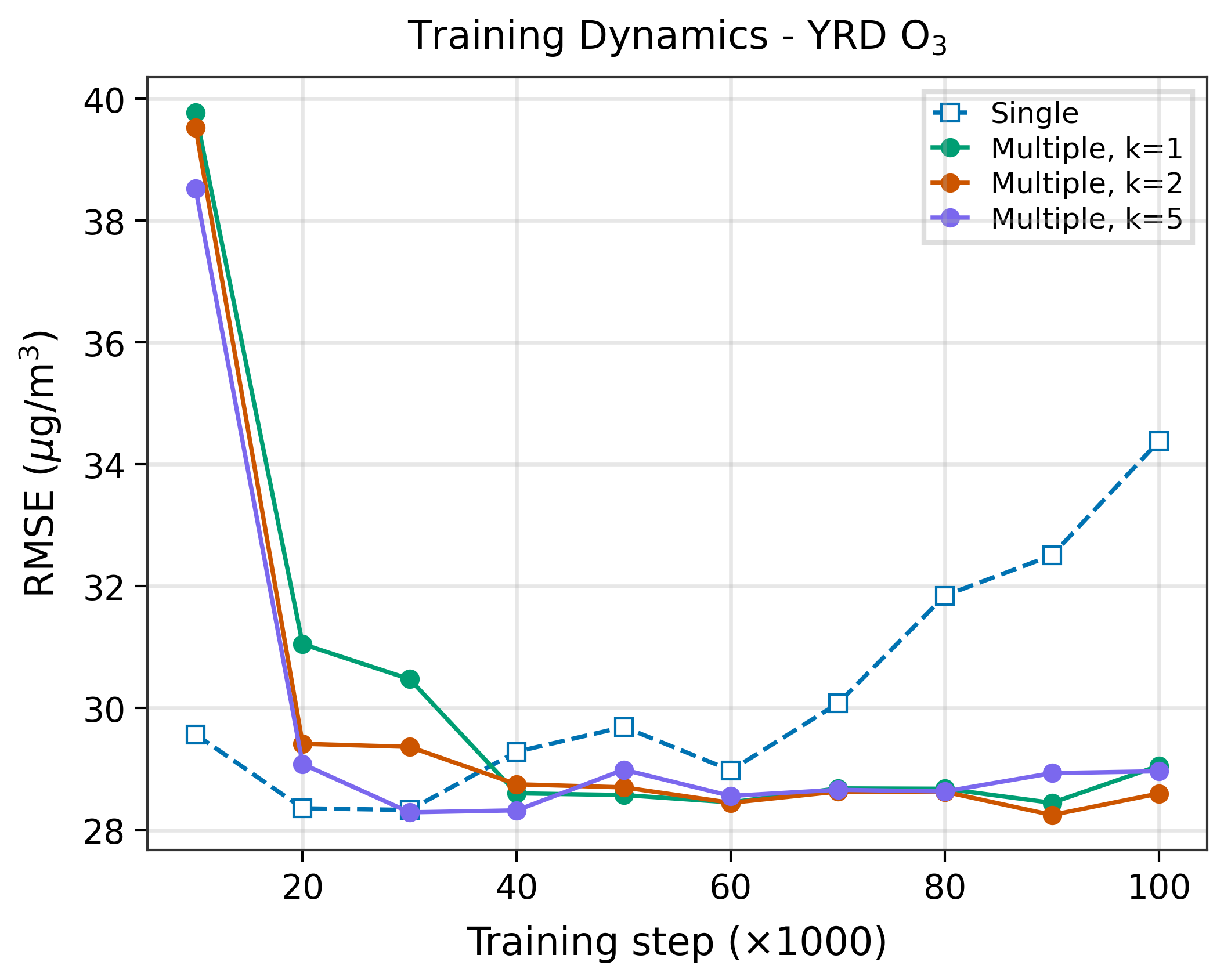}
    \end{minipage}
    \caption{Training dynamics: multi-operator vs.\ single-operator comparison at $\Delta t = 24$h. Left: PM$_{2.5}$. Right: O$_3$. Top: BTHSA. Bottom: YRD. ICON with operator diversity exhibits negligible overfitting while single-operator models overfit.}
    \label{fig:appendix-training-main}
\end{figure}

This section provides training dynamics for both datasets, evaluated at $\Delta t = 24$h. Single-operator models exhibit clear overfitting: validation performance deteriorates despite continued improvement on training loss. In contrast, ICON with operator diversity exhibits negligible overfitting across the entire training process (Figure~\ref{fig:appendix-training-main}). All comparison experiments in this paper use checkpoints at 90,000 training steps to ensure a fair comparison across all models under the same training budget.

\begin{figure}[H]
    \centering
    \begin{minipage}{0.48\textwidth}
        \centering
        \includegraphics[width=\linewidth]{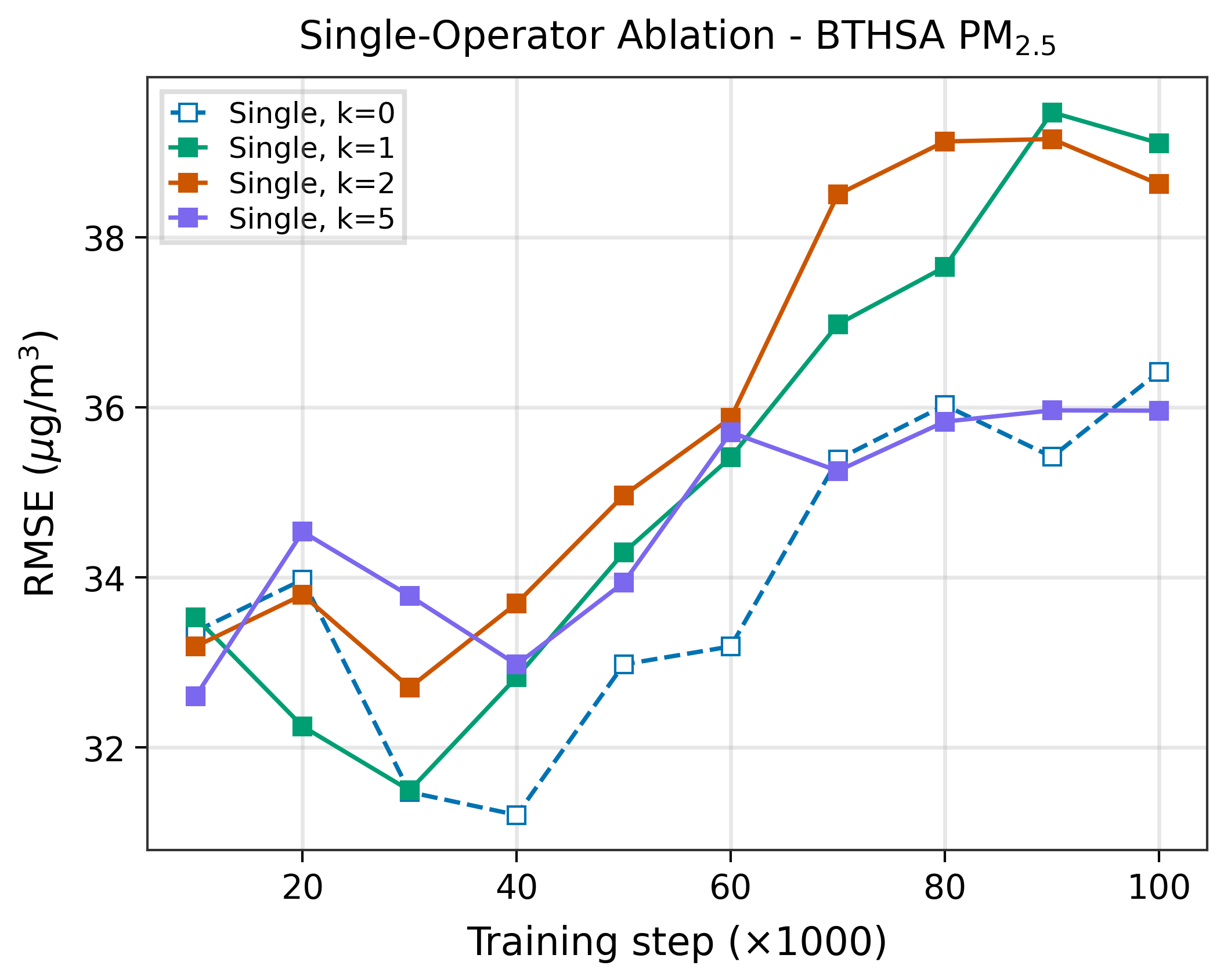}
    \end{minipage}
    \hfill
    \begin{minipage}{0.48\textwidth}
        \centering
        \includegraphics[width=\linewidth]{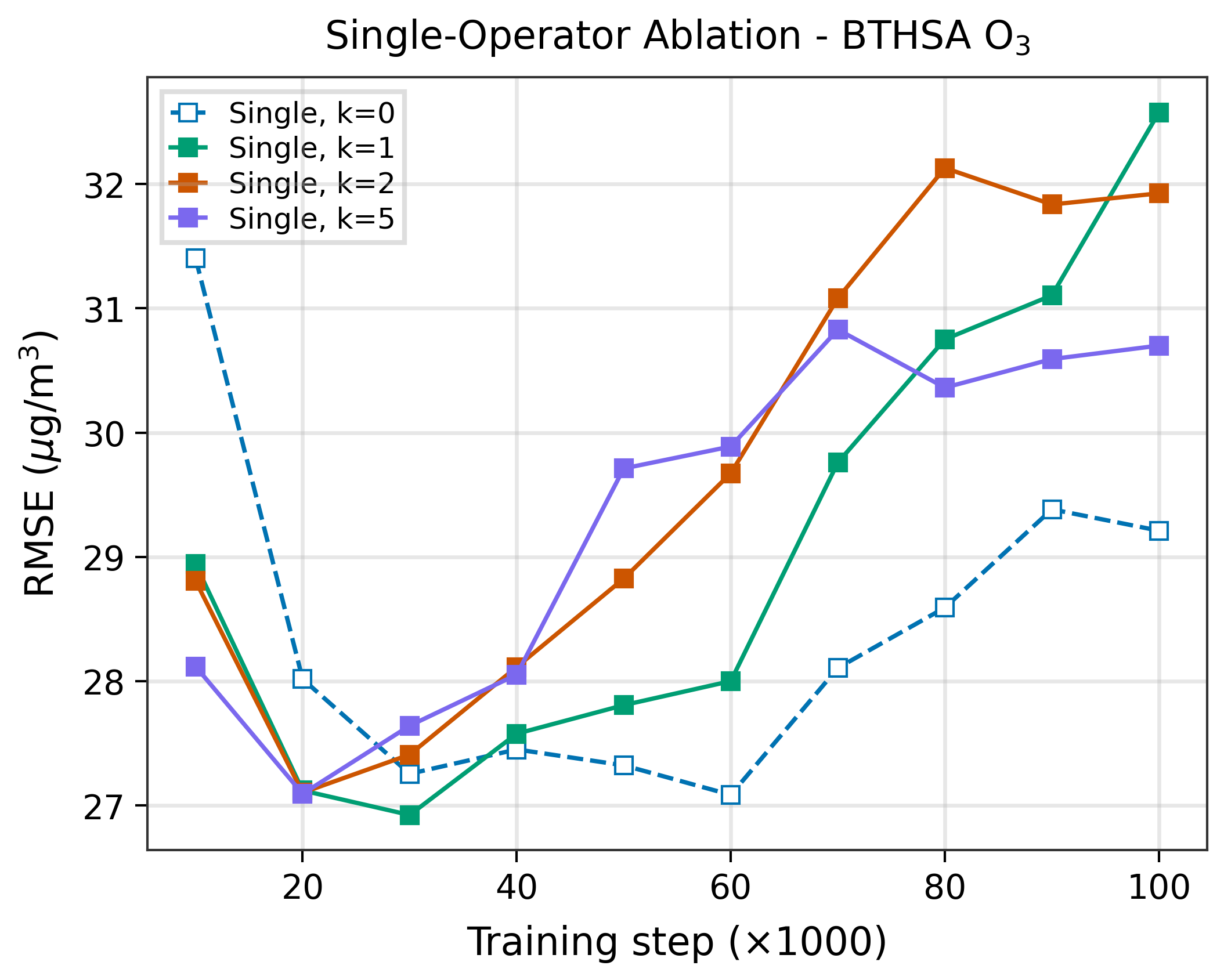}
    \end{minipage}
    \\[0.3em]
    \begin{minipage}{0.48\textwidth}
        \centering
        \includegraphics[width=\linewidth]{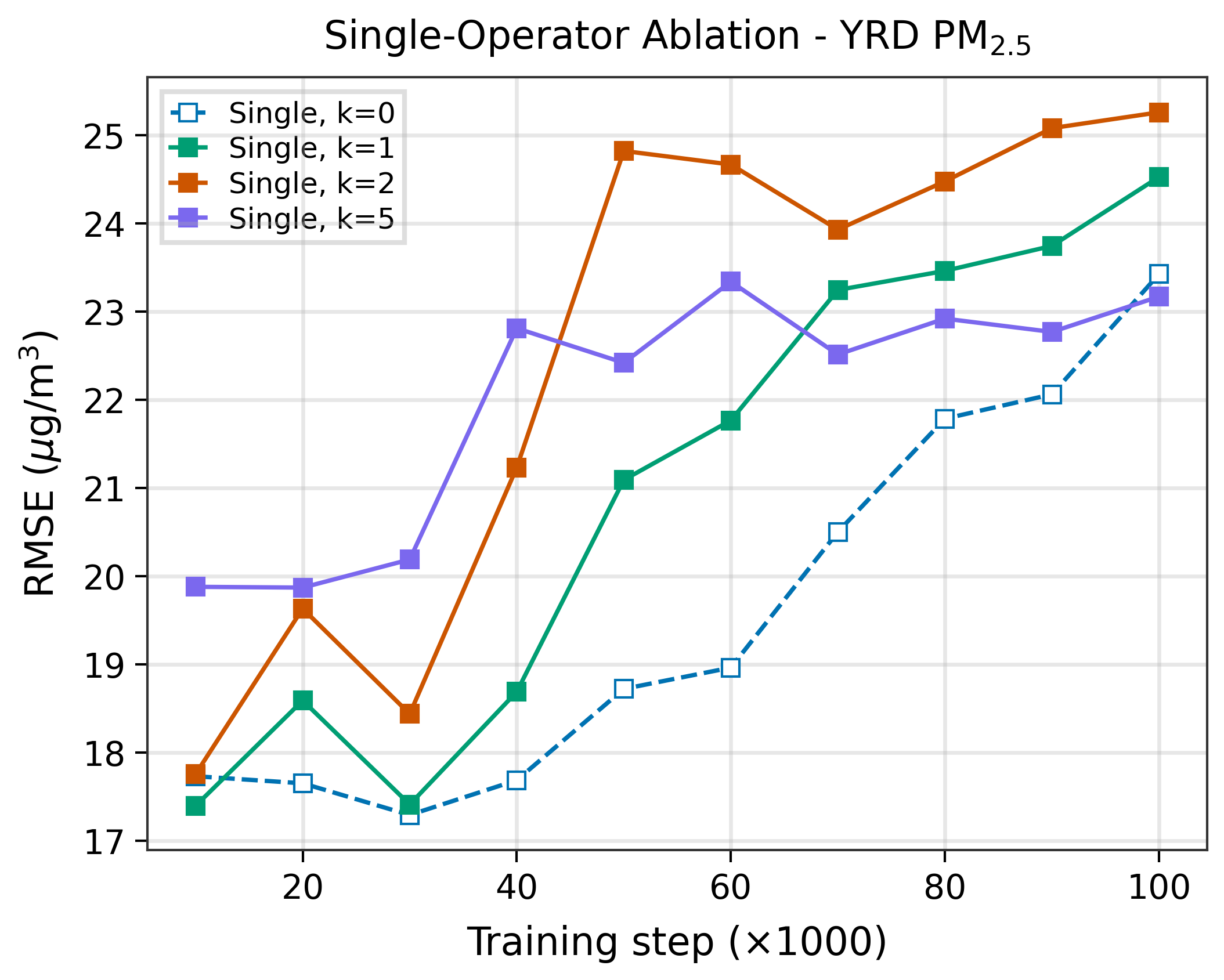}
    \end{minipage}
    \hfill
    \begin{minipage}{0.48\textwidth}
        \centering
        \includegraphics[width=\linewidth]{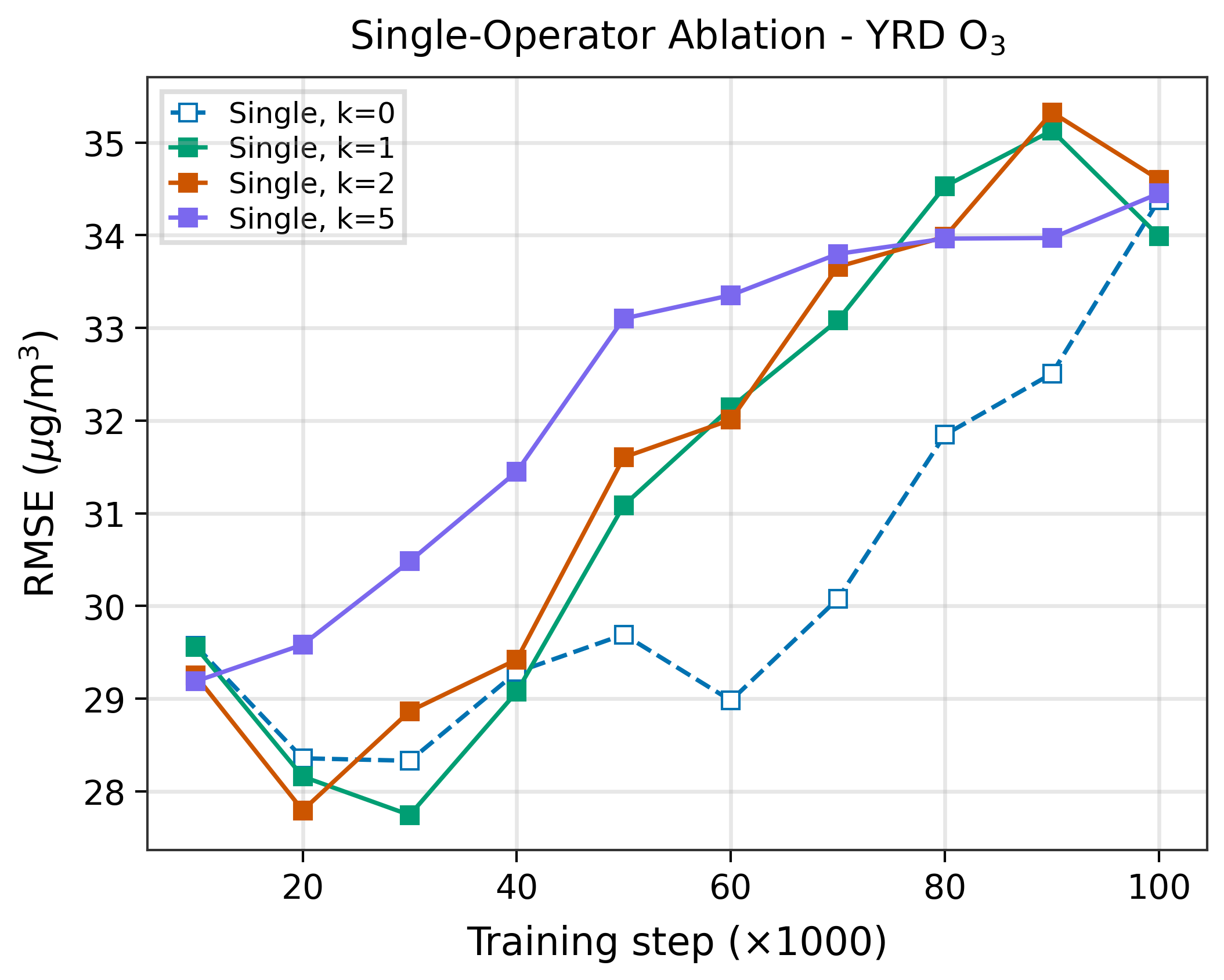}
    \end{minipage}
    \caption{Training dynamics: single-operator ablation at $\Delta t = 24$h with varying $k$. Left: PM$_{2.5}$. Right: O$_3$. Top: BTHSA. Bottom: YRD. All settings exhibit overfitting, with $k = 0$ achieving the best validation performance across all single-operator settings.}
    \label{fig:appendix-training-ablation}
\end{figure}

Figure~\ref{fig:appendix-training-ablation} shows training dynamics for the single-operator ablation with varying maximum example count $k$. All settings exhibit overfitting, with $k = 0$ achieving the best validation performance across all single-operator settings.

\clearpage
\section{Example Cardinality Generalization}
\label{sec:appendix-demo-cardinality}

\begin{figure}[H]
    \centering
    \begin{minipage}{0.48\textwidth}
        \centering
        \includegraphics[width=\linewidth]{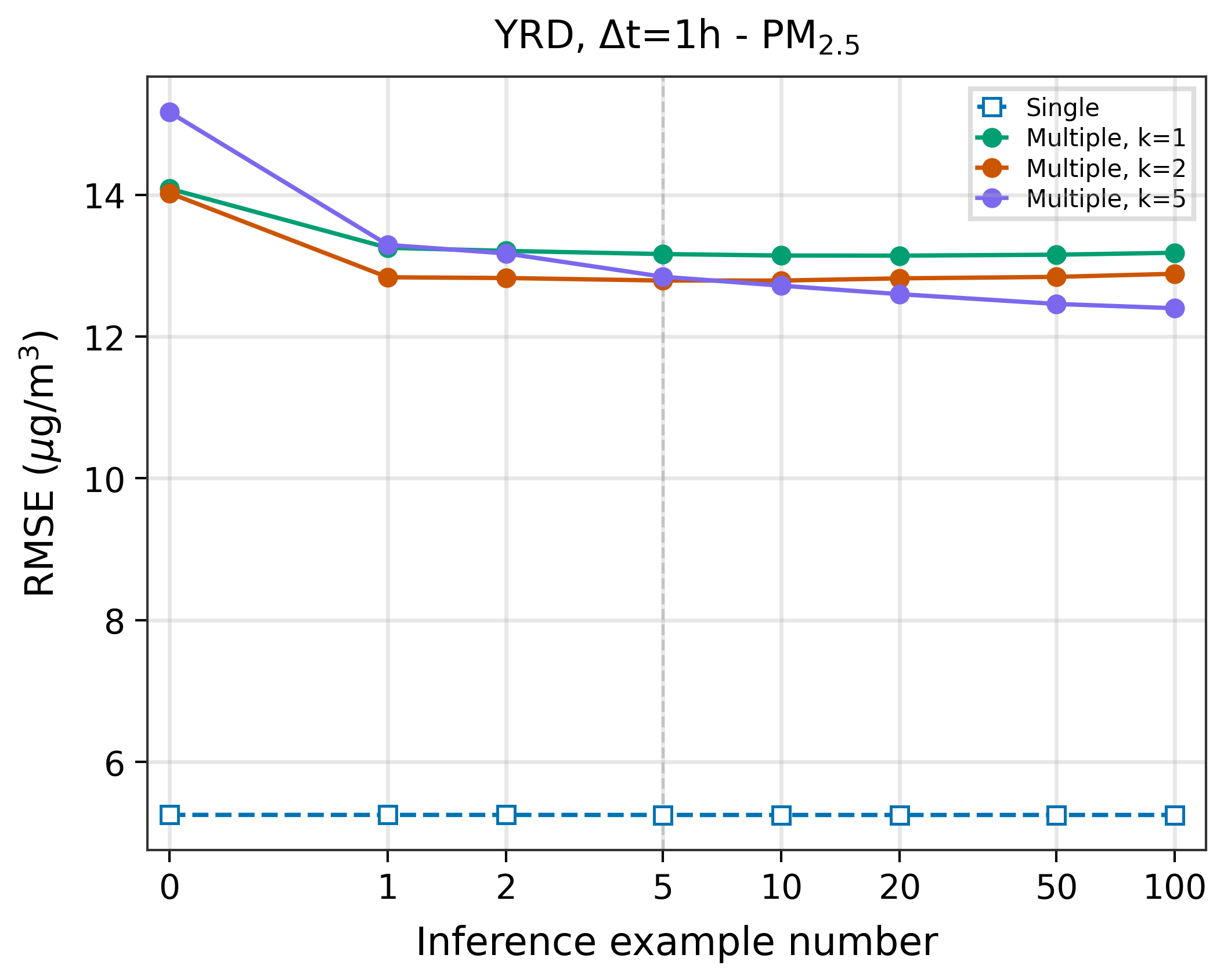}
    \end{minipage}
    \hfill
    \begin{minipage}{0.48\textwidth}
        \centering
        \includegraphics[width=\linewidth]{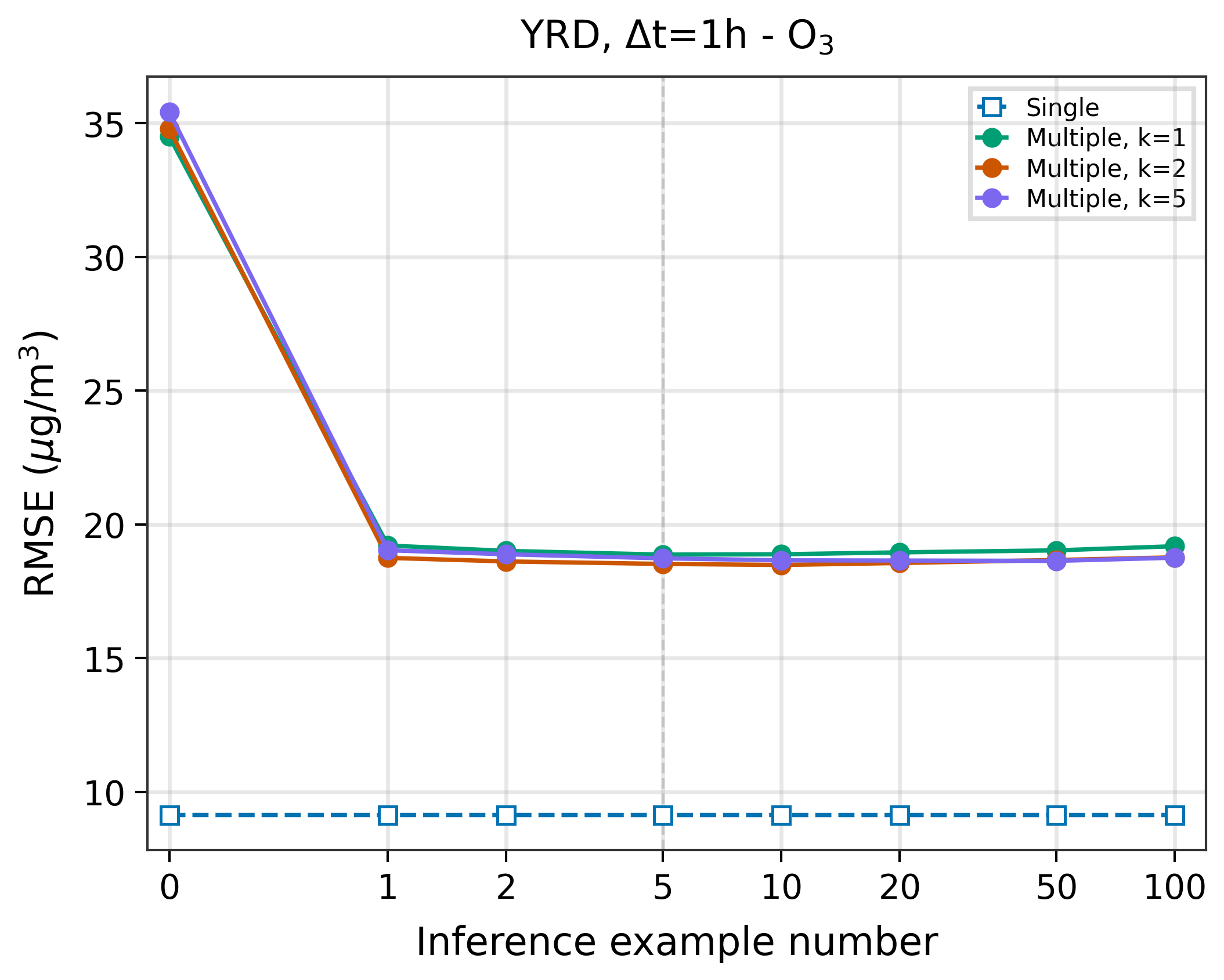}
    \end{minipage}
    \\[0.3em]
    \begin{minipage}{0.48\textwidth}
        \centering
        \includegraphics[width=\linewidth]{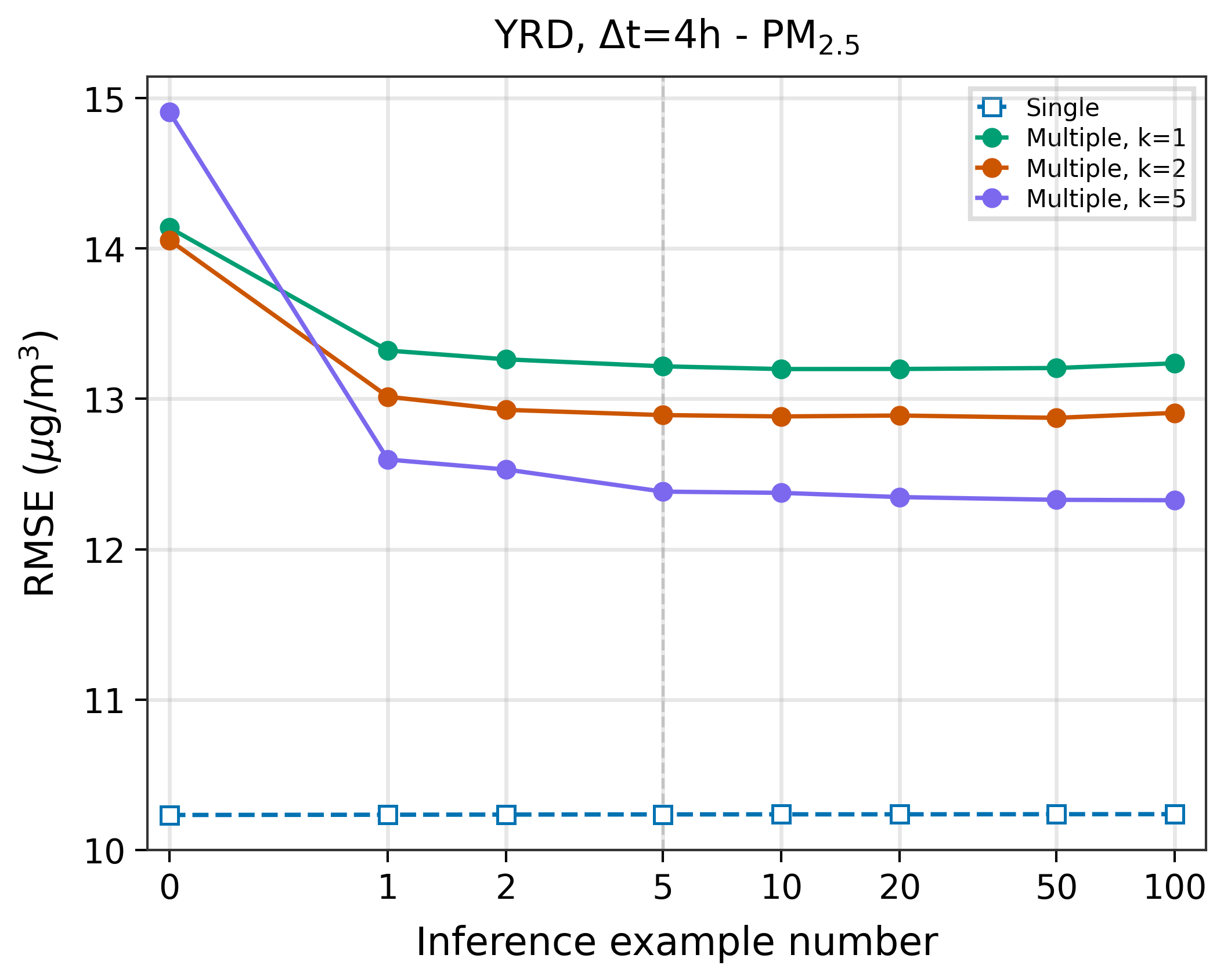}
    \end{minipage}
    \hfill
    \begin{minipage}{0.48\textwidth}
        \centering
        \includegraphics[width=\linewidth]{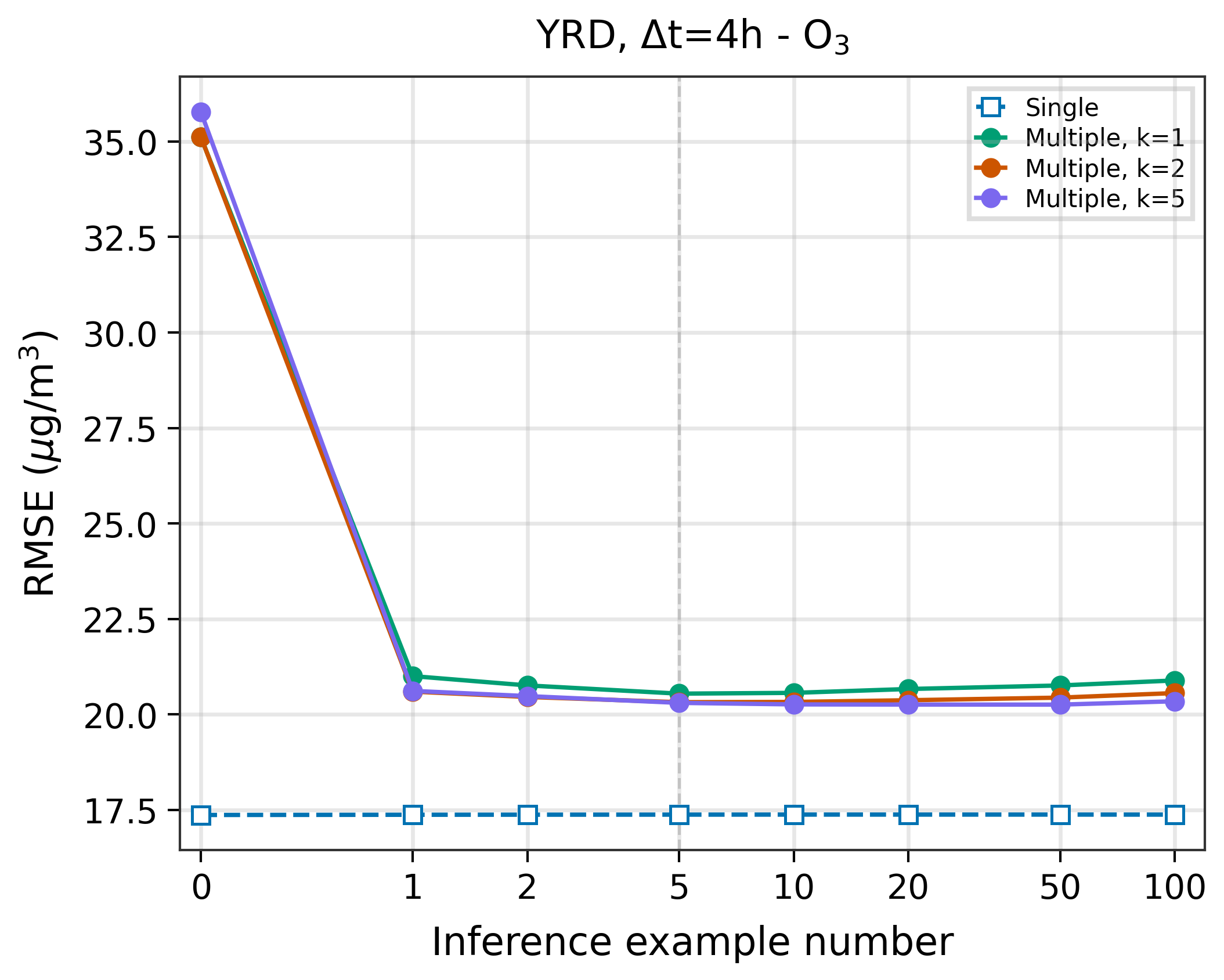}
    \end{minipage}
    \caption{Example cardinality generalization on YRD for simple operators. Left: PM$_{2.5}$. Right: O$_3$. Top: $\Delta t = 1$h. Bottom: $\Delta t = 4$h. Classical single-operator learning achieves lower RMSE for these simple operators.}
    \label{fig:appendix-yrd-simple}
\end{figure}

\begin{figure}[H]
    \centering
    \begin{minipage}{0.48\textwidth}
        \centering
        \includegraphics[width=\linewidth]{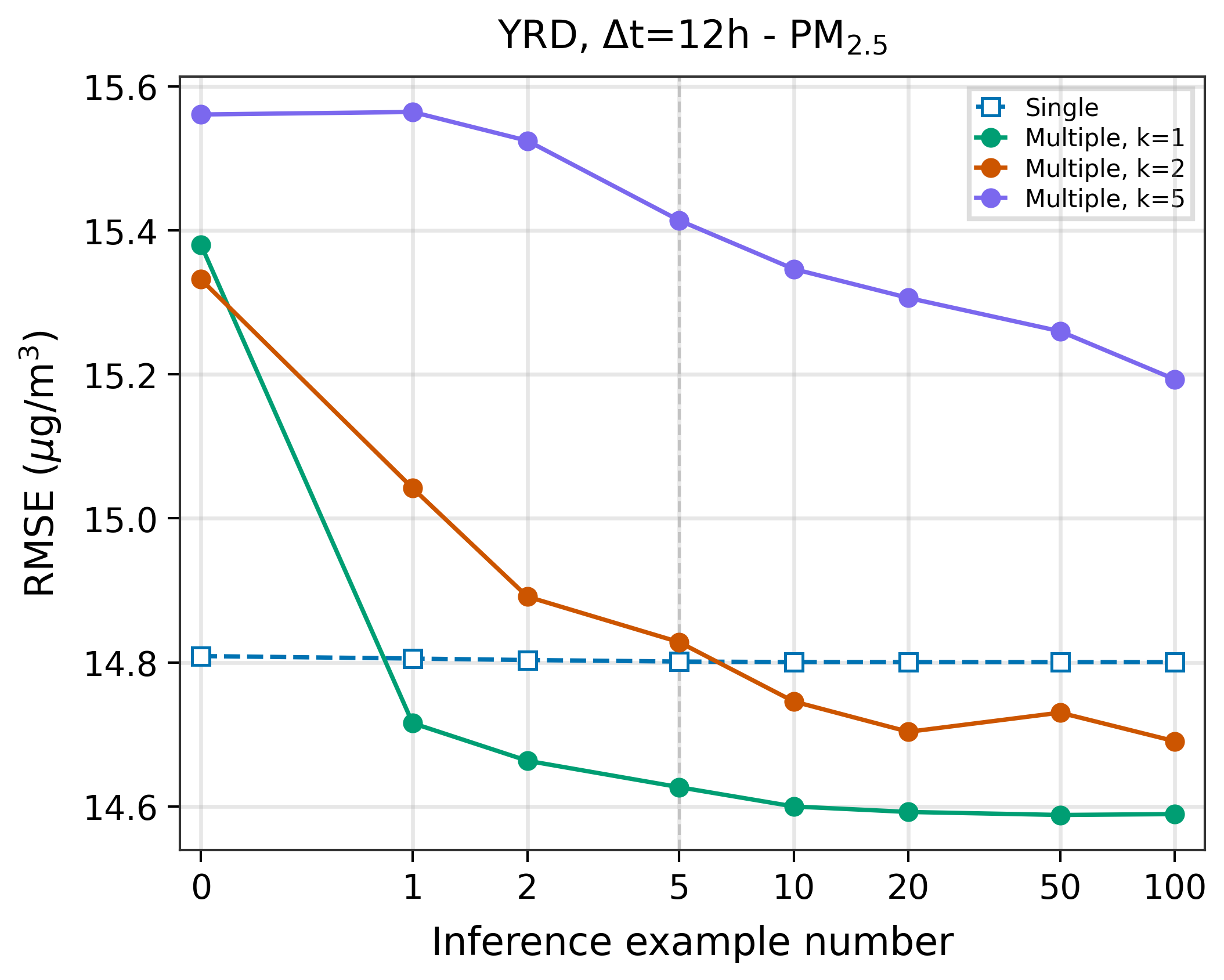}
    \end{minipage}
    \hfill
    \begin{minipage}{0.48\textwidth}
        \centering
        \includegraphics[width=\linewidth]{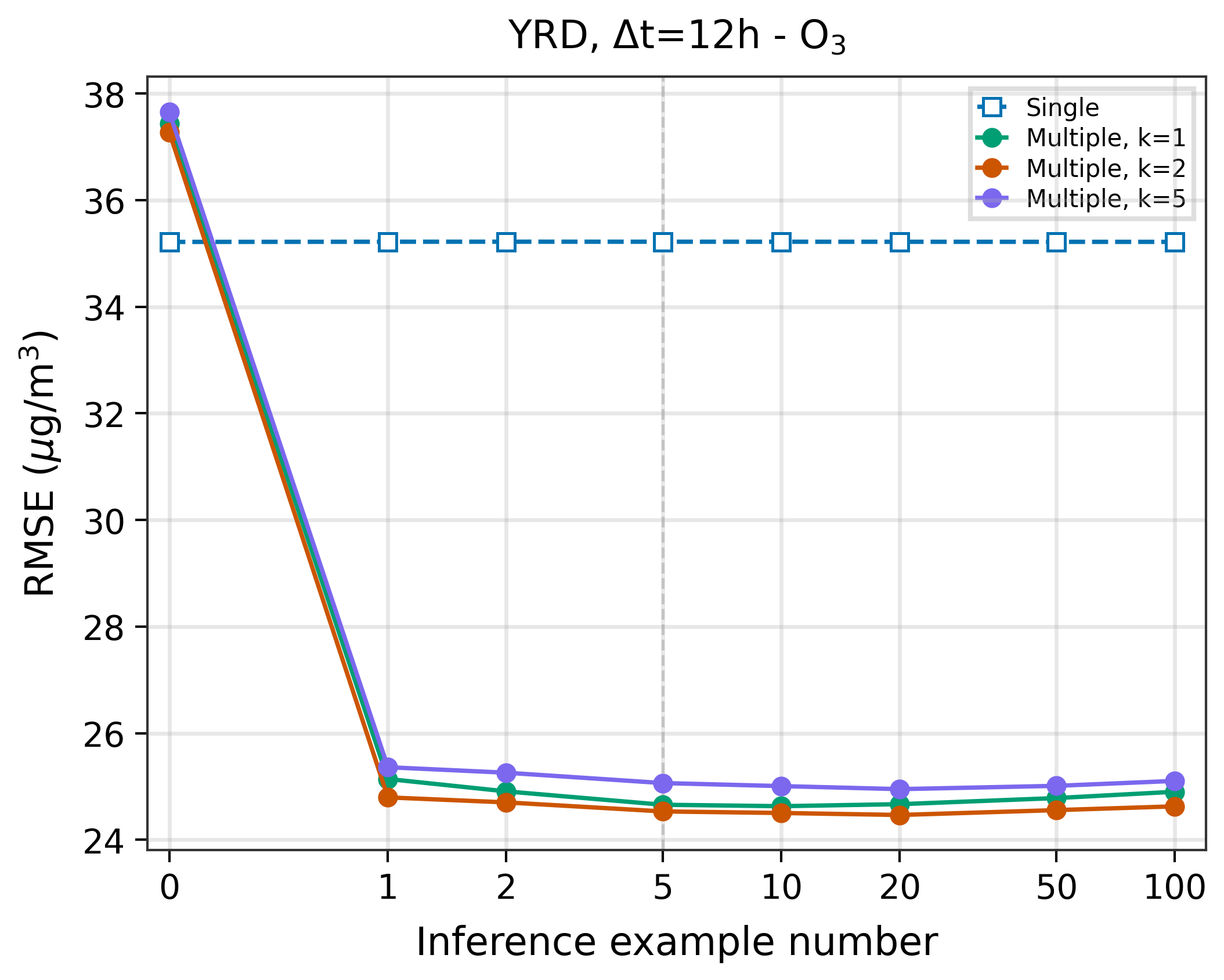}
    \end{minipage}
    \\[0.3em]
    \begin{minipage}{0.48\textwidth}
        \centering
        \includegraphics[width=\linewidth]{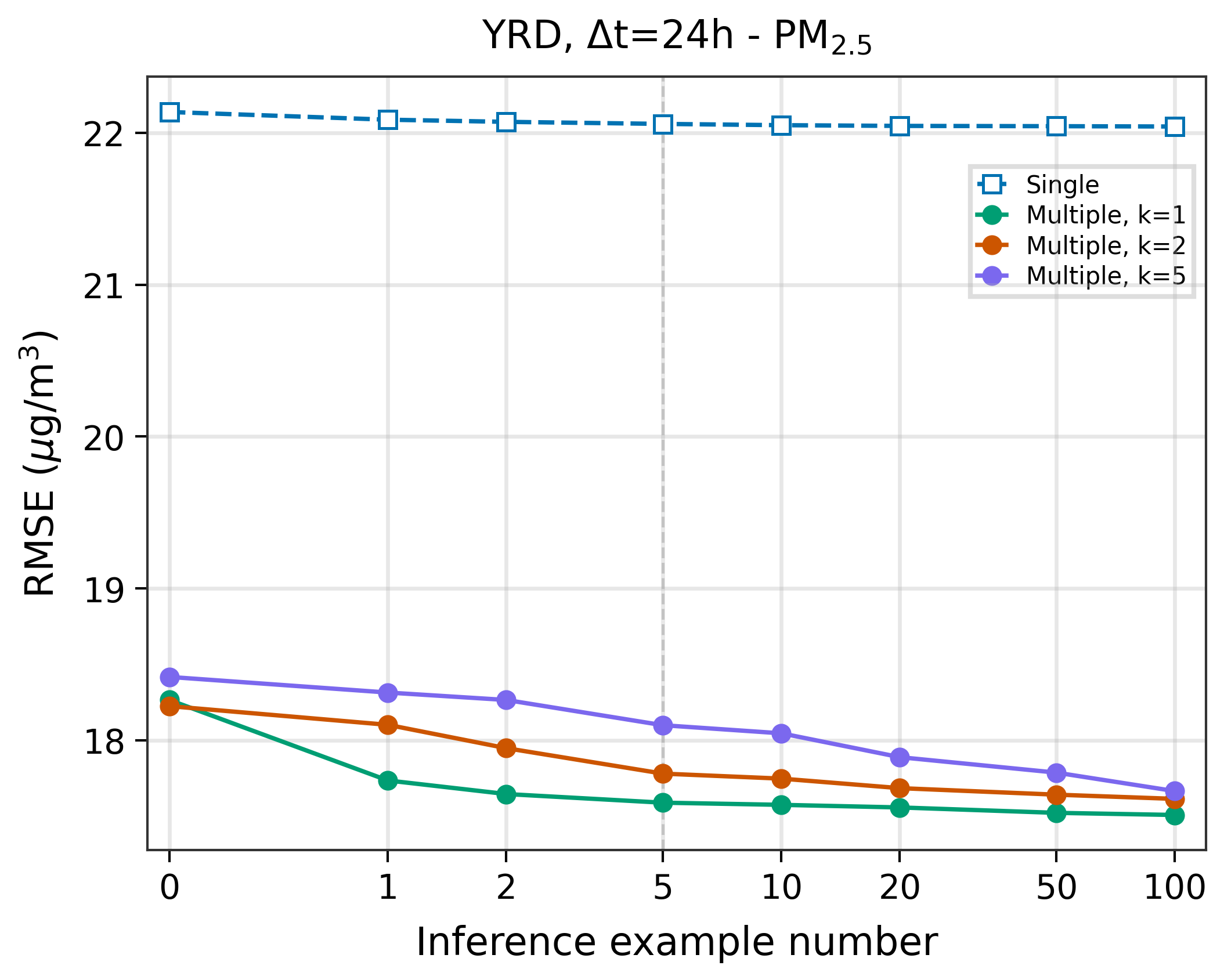}
    \end{minipage}
    \hfill
    \begin{minipage}{0.48\textwidth}
        \centering
        \includegraphics[width=\linewidth]{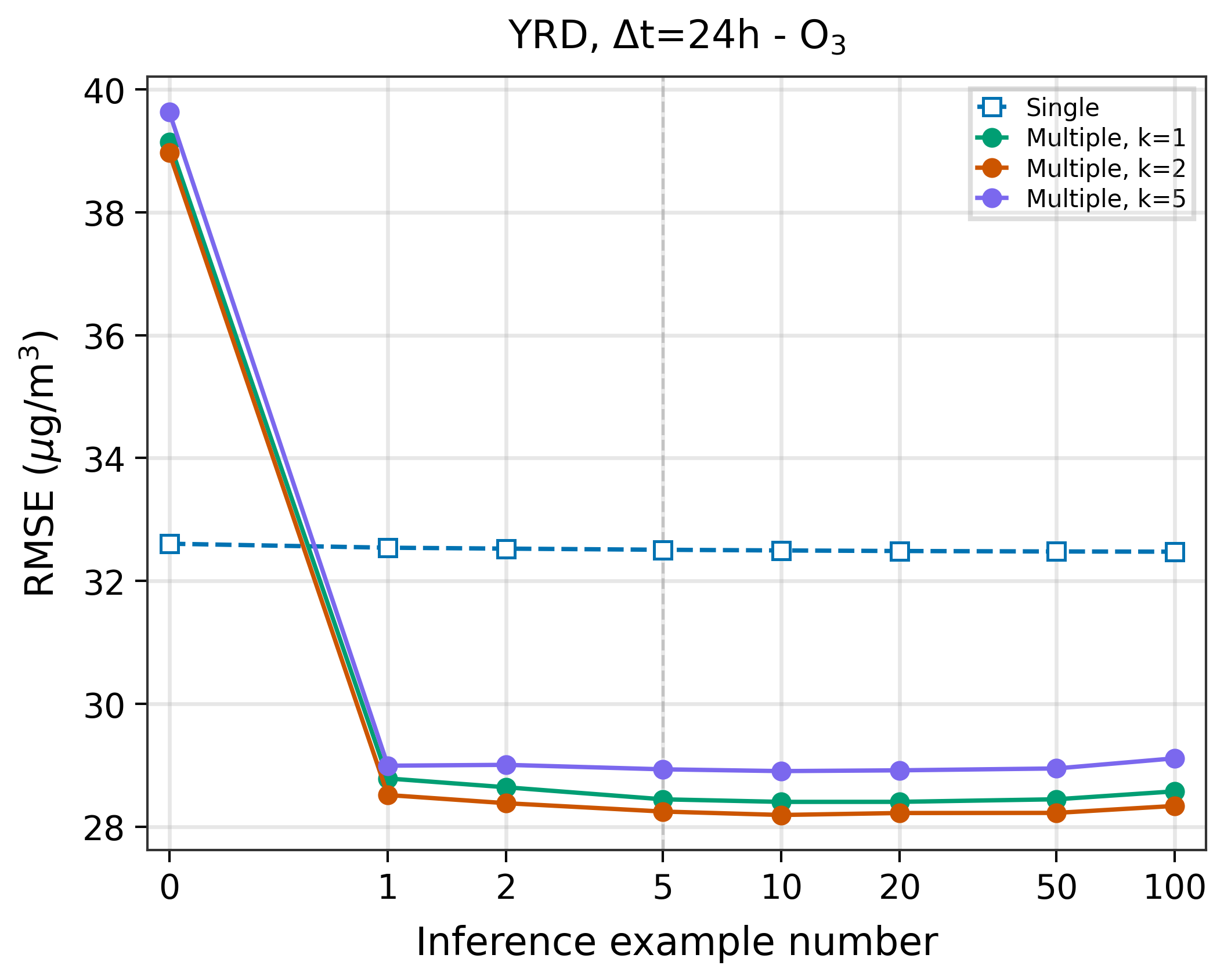}
    \end{minipage}
    \caption{Example cardinality generalization on YRD for complex operators. Left: PM$_{2.5}$. Right: O$_3$. Top: $\Delta t = 12$h. Bottom: $\Delta t = 24$h. ICON with operator diversity surpasses the baseline with sufficient examples.}
    \label{fig:appendix-yrd-complex}
\end{figure}

This section provides example cardinality generalization results for the YRD dataset, complementing the BTHSA results shown in the main text (Figures~\ref{fig:demo_cardinality} and~\ref{fig:demo_cardinality_24}). The same trend is observed: for simple operators ($\Delta t = 1, 4$h), classical single-operator learning achieves lower RMSE (Figure~\ref{fig:appendix-yrd-simple}), while for complex operators ($\Delta t = 12, 24$h), ICON with operator diversity surpasses the baseline with sufficient examples (Figure~\ref{fig:appendix-yrd-complex}).

\end{document}